\documentclass[journal, A4]{IEEEtran} 
\usepackage{amsmath,amssymb,amsfonts}
\usepackage{algorithmic}
\usepackage{algorithm}
\usepackage{array}
\usepackage[caption=false,font=normalsize,labelfont=sf,textfont=sf]{subfig}
\usepackage{textcomp}
\usepackage{stfloats}
\usepackage{url}
\usepackage{verbatim}
\usepackage{graphicx}
\usepackage{xcolor}
\begin{document}

\title{Thermal and Visual Tracking of Photovoltaic Plants for Autonomous UAV Inspection}
%
%

\author{Luca Morando, Carmine Tommaso Recchiuto, Jacopo Call\`a, Paolo Scuteri and Antonio Sgorbissa\thanks{L. Morando, C. Recchiuto, A. Sgorbissa are with University of Genova; J. Calla and P. Scuteri are with JPDroni S.r.l. // Contact author email: \texttt{antonio.sgorbissa@unige.it}}
}%




\maketitle

	\begin{abstract}
Since photovoltaic (PV) plants require periodic maintenance, using Unmanned Aerial Vehicles (UAV) for inspections can help reduce costs. Usually, the thermal and visual inspection of PV installations works as follows. A UAV equipped with a Global Positioning System (GPS) receiver is assigned a flight zone, which the UAV will cover back and forth to collect images to be later composed in an orthomosaic. When doing this, The UAV typically flies at a height above the ground that is appropriate to ensure that images overlap even in the presence of GPS positioning errors. However, this approach has two limitations. Firstly, it requires covering the whole flight zone, including ``empty" areas between PV module rows. Secondly, flying high above the ground limits the resolution of the images to be later inspected.  
The article proposes a novel approach using an autonomous UAV with an RGB and a thermal camera for PV module tracking. The UAV moves along PV module rows at a lower height than usual and inspects them back and forth in a boustrophedon way by ignoring ``empty" areas with no PV modules. Experimental tests performed in simulation and an actual PV plant are reported.
	\end{abstract}

\begin{IEEEkeywords}
Aerial Systems, Autonomous Vehicle Navigation, Photovoltaic (PV) plant inspection
\end{IEEEkeywords}

\section{Introduction}
\begin{figure}[t!]
\centering
\includegraphics[width=0.9 \linewidth]{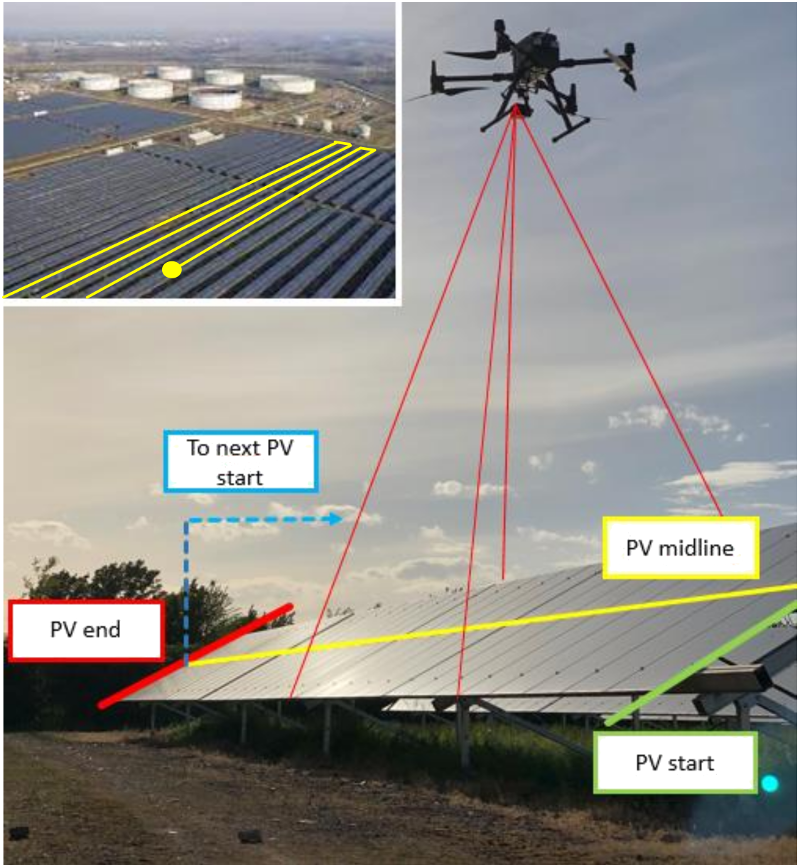}
	\caption{System at work in a PV plant. The DJI Matrice 300 drone was equipped with a hybrid RGB and a thermal camera, the DJI Zenmuse XT2. 
}
\label{fig:figure1}
\end{figure}

\IEEEPARstart{W}{e} are currently facing a worldwide energy challenge that requires us to search for alternatives to fossil fuels, including green and renewable energies \cite{Heinberg}.
According to \cite{EU_website}, in 2019, renewable energy sources made up 34\% of gross electricity consumption in the EU-27, slightly up from 32\% in 2018. While wind and hydropower accounted for two-thirds of the total electricity generated from renewable sources (35\% each), the remaining one-third of the electricity generated was from solar power (13\%), solid biofuels (8\%), and other renewable sources (9\%). The analysis also shows how solar power is the renewable source experiencing the fastest growth, given that in 2008 it accounted for around 1\%. 

Solar energy plants offer many advantages since they have a long life, are environmentally friendly, noise-free, and clean. However, photovoltaic (PV) installations need periodic maintenance since they always need optimal conditions to work properly \cite{Petrone}.
Surface defects \cite{GRIMACCIA, Tsanakas, Quater, Carletti, Djordjevic} are the most common problems. They can be detected through human inspection: a qualified operator can easily detect different defects, including snail trails\footnote{A snail trail is a discoloration of the panel.}, yellowing of the encapsulant, delamination and bubble formation in the encapsulant, front surface soiling, 
busbar oxidation, or impact from physical objects. However, a human inspection might be time-consuming if the PV plant is very large (or in particular conditions, e.g., panels are mounted on a rooftop). 
In order to reduce the cost and time required for maintenance, methods exist \cite{line_tracking_PV_detection} to estimate the presence and impact of global defects through the analysis of the power output.
However, this approach presents two main disadvantages: a reference power production is required, and the exact locations of defects cannot be identified.   

Unmanned Aerial Vehicles (UAVs) have been recently proposed for PV inspections. In the past decades, research made significant steps forward concerning the development of UAVs for monitoring applications, including the inspection of power transmission lines \cite{Hartmut}, gas and oil pipelines \cite{Rathinam}, precision agriculture  \cite{Honkavaara}, and bridges \cite{METNI20073}.  
Indeed, the ability of multirotor UAVs to hover and move freely in the air and the fact that they can be easily piloted and equipped with different sensors make this technology very appealing in monitoring scenarios. Generally, UAVs used for PV inspection are equipped with a thermal camera (which may or may not complement a standard RGB camera or other sensors) to identify defects that can produce heat anomalies on the solar panels.

The use of UAVs reduces the inspection time but, on the other hand, creates a larger amount of raw data to be processed, which must meet some requirements in terms of resolution and position accuracy. 
Currently, the inspection of PV plants through thermovisual imagery is mostly based on UAV photogrammetry \cite{Aghaei2016, Zefri2018, Zefri2022}. A UAV equipped with a Global Positioning System (GPS) receiver and an Inertial  Measurement Unit (IMU) is assigned a flight zone: it will cover this area in a ``boustrophedon" way (i.e., back and forth like the ox turns while plowing) by ensuring the required overlapping of images, horizontally and vertically, to be later composed in an orthomosaic. The UAV  typically flies at a height above the ground that is appropriate to ensure that the images overlap even in the presence of GPS positioning errors. However, this approach has two limitations.   Firstly, it requires covering the whole zone including ``empty" areas between PV module rows: depending on how PV module rows have been arranged in the PV plant, there may be large areas without PV panels to be inspected. Secondly, flying high above the ground limits the impact of positioning errors, which is good, but also the resolution of the images to be later inspected.

One could imagine a solution that does not build an orthomosaic of the whole plant. The UAV can be instructed to move along PV module rows at a lower height, Figure \ref{fig:figure1}, so that the panel almost completely occupies the camera's field of view and ignores the areas where there are no PV modules to be inspected. This process will produce ``strips" of higher resolution images (one strip per PV module row) instead of a ``mosaic." To implement this mechanism, it might seem sufficient to provide a sequence of waypoint coordinates to the UAV's mission planner  before take-off. The waypoints might, in their turn, be chosen during the pre-flight setup, with an operator drawing the desired path on Google Earth images showing the solar plant to be inspected. However, as discussed in \cite{line_tracking_PV_detection}, the error in planimetric coordinates of the objects captured in Google Earth 
ranges from 5 to 10 meters. This aspect, together with the intrinsic inaccuracy of the GPS signal, will likely produce a wrong UAV placement, determine a wrong alignment of the UAV with the PV module rows, increase the amount of useless data acquired, and reduce performances.

Based on these premises, a different solution using sensor information to correct in real-time the error between the UAV and the actual position of PV modules is needed. 
The main contribution of this article is a portfolio of techniques for PV module segmentation and UAV navigation through visual servoing based on the onboard RGB and thermal cameras. Please remark that, in PV plant inspection, a thermal camera is required for detecting defects: then, using both RGB and thermal cameras for PV module row tracking is particularly convenient as it may improve reliability in critical light or temperature conditions \cite{line_tracking_PV_detection}. JP Droni\footnote{JP Droni Srl is a company in Genova providing aerial services for video productions, precision agriculture, and technical inspection of power plants.} confirms that  GPS-based photogrammetry is the only approach currently adopted in Italy (and, to the best of their knowledge, in the world) for PV inspection: in commercial applications, the UAV typically flies at 30-40 meters above the ground. With respect to GPS-based photogrammetry, the availability of new visual servoing techniques based on PV module segmentation might produce two breakthroughs. First, it allows the drone to follow a more accurate navigation direction along the planned path, which lies in the middle of the underlying PV array to detect. Thanks to higher accuracy in drone localization with respect to the PV array, the system provides a solution to the wrong placement of the GPS waypoints (e.g., done through Google Earth before the mission's start): this, in its turn, prevents the drone from flying over empty areas between two parallel PV arrays, collecting useless data and wasting time and battery autonomy. Second, it enables the drone to fly at a lower height to the ground, capturing details on the PV array surfaces otherwise impossible to see (including PV panels' serial numbers) while reducing the oscillations in position generated by noise in GPS localization.    

Figure \ref{fig:figure1} shows how the system works and introduces the main concepts used in the rest of the article. 

\begin{itemize}
	\item \textit{PV midline}, a straight line in the middle of the PV module row that determines the desired motion direction.
	\item \textit{PV end}, a point on the PV midline that identifies the end of the PV module row. 
	\item \textit{PV start}, a point that identifies the start of the new PV module row, whose position is computed with respect to the end of the previous row. 
\end{itemize}
The upper left corner of Figure \ref{fig:figure1} shows a UAV moving along the PV rows in a boustrophedon way. The UAV moves from \textit{PV start} to \textit{PV end} along a \textit{PV midline}. Then, it ``jumps" to the next PV row and it starts moving again from the following \textit{PV start} to its corresponding \textit{PV end}, and so on.

The article is organized as follows. Section \ref{State of the Art} surveys the relevant literature.  Section \ref{sec: System Architecture} describes the system architecture. Section \ref{sec: Panel's arrays segmentation} introduces the techniques used for PV module segmentation using RGB and thermal images. 
Section \ref{sec: UAV Navigation} introduces strategies for UAV  autonomous navigation. Section \ref{Experiments} presents the experimental results in a photorealistic simulated environment and a real PV plant. Finally, in Section \ref{Conclusion}, conclusions and directives for future research are given.

\section{State of the Art}
\label{State of the Art}


Various techniques and applications have been proposed for automatic defect detection from aerial images. For example, the authors of  \cite{Hassan2020} and \cite{Pan2018} use computer vision and machine learning techniques to detect and classify cracks and potholes in roads and highways using images taken by UAVs. 
Similarly, \cite{Zhang2019} and \cite{Zormpas2018} address the inspection of power lines with UAVs, using techniques such as Faster region-based convolutional neural network (Faster R-CNN) \cite{ren2016faster} to detect defects and the Hough Transform \cite{Duda197211} for cable detection. 
Strategies for automatic defect detection have also been applied to the PV plant scenario. For example, \cite{Tsanakas2017}  proposes two different techniques for inspection and mapping, aerial IR/visual image triangulation and terrestrial IR/visual image georeferencing: the authors discuss the potential of both approaches in localizing defective PV modules. 
In \cite{Xiaoxia2019}, a Convolutional neural network (CNN) is used for defect recognition based on aerial images obtained from UAVs. The authors of \cite{Huerta2020} present an approach that unifies two region-based CNNs (R-CNN) to generate a robust detection approach combining thermography and telemetry data for panel condition monitoring.  
A model-based approach for the detection of panels is proposed in \cite{Carletti2020AnIF}: this work relies on the structural regularity of the PV arrays and introduces a novel technique for local hot spot detection from thermal images, based on a fast and effective algorithm for finding local maxima in the PV panel regions. In \cite{Fernandez2020} a fully automated approach for detecting, classifying, and geopositioning thermal defects in the PV modules is proposed, which exploits a novel method based on image reprojection and function minimization to detect the panels from the images.  The work described in \cite{Chiang2022} proposes a novel method for estimating the power efficiency of a PV plant using data from thermal imaging and weather instruments obtained using an UAV instrumented with a radiometer, a thermometer and an anemometer. These and similar works address only the problem of autonomous defect detection and geopositioning without using the acquired images as feedback for the UAV to move autonomously along the inspection path.

The autonomous inspection of PV plants through UAV photogrammetry has been explored in the literature \cite{Aghaei2016, Zefri2018, Zefri2017,  Souffer2022}. The UAV is given a set of waypoints, usually arranged in such a way to cover a delimited area to ensure the required horizontal and vertical overlapping of images. 
Then, the UAV moves along the planned sequence of waypoints using GNSS/GPS data only, possibly negatively impacting data quality due to positioning errors \cite{Solend2021}. 
An approach to generating an optimal waypoint path based on satellite images is proposed in \cite{Moradi2019} to reduce the error due to the misalignment of georeferenced images and the actual inspection site. The PV plant's boundary is extracted via computer vision techniques, and the system can re-compute the path if the UAV requires reaching a specific location (e.g., a PV module where a defect has been detected) or if the UAV cannot complete the initial path for any reason. In \cite{Moradi2020}, a novel technique for boundary extraction is proposed using a Mask R-CNN architecture with a modified VGG16 backbone. The network is trained on the AMIR dataset \cite{amir_dataset}: an aerial imagery collection of PV plants from different countries. The authors of \cite{Perez-Gonzalez2021} present a novel way to develop a flight path automatically with coverage path planning (CPP) methods \cite{Le2022} by segmenting the region of interest within PV plant images. 
The work proposed in \cite{Zefri2022} uses a VGG16 CNN fed with images that underwent a Structure from Motion – MultiView Stereo (SfM-MVS) photogrammetric processing workflow to generate thermal orthomosaics of the inspected sites.  These and similar approaches focus on planning a waypoint path with the objective to cover the whole area according to the principles of photogrammetry, i.e., they do not consider tracking individual PV module rows at a lower height from the ground to produce fewer and higher resolution images to be processed. 

Visual Servoing is widely used in robotics. Many examples of UAV autonomous navigation based on data acquired through cameras exist in the literature, some involving inspection tasks.  For example, an algorithm for detecting vertical features from images has been used in \cite{Rathinam2005} to guide a UAV along a semi-linear path while flying over a highway in the US. This approach is based on the consideration that vertical edges are a characterizing attribute of highways that can be easily detected using methods such as RANSAC \cite{Ransac}, which can also be used to predict splines \cite{Majeed2020} where sharp curves may occur. 
Similar techniques may be found in \cite{Sarapura2017} and \cite{Li2019}, where the authors respectively contributed to the field of precision agriculture and building inspection. 
In \cite{Sarapura2017}, a novel control technique based on a passivity-based visual servoing controller with dynamics compensation to track crops is described. The proposed system allows a UAV to move along straight lines such as those formed by structured crops. 
In \cite{Li2019},  a complete navigation and inspection task is executed by an autonomous quadrotor. The most interesting part of this work is the approach used for road detection and navigation, which segments the image into predominant color layers before feeding it to DeepLabv3+  \cite{Chen2018}, a network trained for semantic segmentation. The road is then identified in the image and finally followed by the UAV by defining a sequence of control points. 
The approach recently proposed in \cite{Shahoud2022} relies on the availability of well-described streets in urban environments for UAV navigation and path tracking: while the drone position is continuously computed using visual odometry, scene matching is used to correct the position drift.

Visual servoing was rarely applied to UAV navigation over large PV plants. This may also be due to the complexity of the scenario, which may have different characteristics in terms of the number of PV module rows, their dimension, and mutual arrangement. 
Nevertheless, some attempts in this direction have been made in recent years. For example, in \cite{Roggi2020}, the authors propose a vision-based guidance law to track PV module rows based on PV modules' edge detection. The approach extracts vertical edges from RGB images using a Hough Transform. Then, the detected misalignment between the PV row observed from images and the actual UAV position is fed to the navigation control law.  
Unfortunately, this approach has only been tested in simulation. 
An edge detection technique is also proposed in \cite{Zhipeng2018}, where the metallic profile of the PV array is detected and segmented by combining color and feature extraction, using the Hough Transform and a Hue-Saturation-Value (HSV) model. Once the panels have been segmented, the velocity orthogonal to the PV module row is controlled to keep the UAV aligned. 
Both approaches rely on the tuning of parameters for edge extraction: 
notwithstanding this limitation, it is worth noticing that, in the ``era of Deep Learning," many of the approaches mentioned above  \cite{Sarapura2017, Li2019, Roggi2020,  Zhipeng2018} rely on model-based vision techniques for feature segmentation. In structured environments whose model is a priori known, such as PV plants, a model-based formulation of the problem may offer advantages in predictable and explainable behaviour \cite{BarredoArrieta2020} and possibly robustness to adversarial attacks \cite{Thys2019} that cannot be ignored. As an aside, this is also confirmed by AI in aviation (particularly Machine Learning) still being at the beginning. Even if AI is widespread in subdomains like logistics and fuel consumption estimation  \cite{Haponik2021, AI}, AI techniques in flight control can hardly be found in real-world application scenarios.

\section{System Architecture}
\label{sec: System Architecture}


The system can autonomously perform UAV-based PV inspection centered on the following core elements.

\begin{itemize}
	\item A procedure for detecting PV modules in real-time using a thermal or an RGB camera (or both).
	\item A procedure for correcting errors in the relative position of the \textit{PV midline}, initially estimated through GPS, by merging thermal and RGB data.
       \item A navigation system provided with a sequence of georeferenced waypoints defining an inspection path over the PV plant, which uses the  estimated position of the \textit{PV midline} for making the UAV move along the path through visual servoing. 
	\end{itemize}

\begin{figure}[t]
	\centering
\includegraphics[width= 0.9\columnwidth]{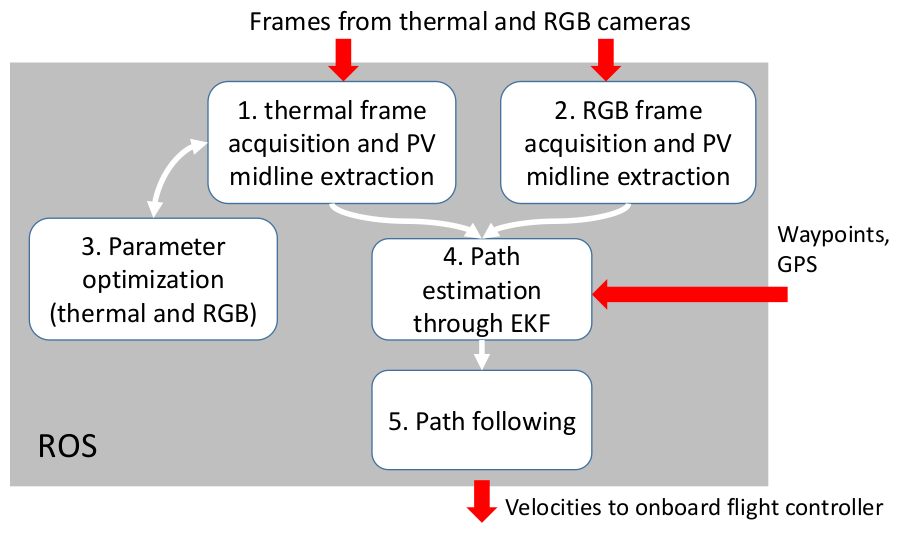}
   \caption{System architecture implemented in ROS.}
    \label{fig:Figure2}
\end{figure}

The software architecture implemented in ROS \cite{ros} is shown in Figure \ref{fig:Figure2}. 
ROS nodes (1) and (2) acquire and process raw frames from the RGB and the thermal camera and perform PV module segmentation, Sections \ref{sec:Detection of PV arrays via thermal camera}, \ref{sec: Detection of PV arrays via RGB camera}. As an output of this process, each node returns the parameters, in the image frame, of the straight lines running in the middle of the detected PV module rows. 
To increase segmentation accuracy with varying environmental conditions, node (3) implements an optimization procedure to find the best parameters for segmentation: 
some details about this procedure are in Sections \ref{sec: Detection Parameters Optimization}.
Node (4) merges subsequent observations, independently performed by the two cameras in subsequent time instants. Since the parameters of straight lines independently extracted by (1) and (2) may be affected by errors, 
an Extended Kalman Filter (EKF) is used to estimate the \textit{PV midline} by merging the observations provided in real-time by nodes (1) and (2) with a priori knowledge, 
Sections \ref{sec: Export of the Control Points from the Image Frame I' to the Body Frame B} and \ref{sec: Estimation of the PV array midline}. 
Finally, node (5) uses the estimated reference path for controlling the UAV along its path through visual servoing, Section \ref{sec: UAV Navigation}.

 PV module row tracking is not sufficient: node (5) also needs rules to instruct the UAV to move to the next row when the previous one has been completed. This means we need a waypoint-based reference path defining the order according to which PV module rows shall be inspected. Remember that waypoints can be labelled as \textit{PV start} or \textit{PV end} and are acquired before the inspection through Google Earth or other georeferenced images (possibly affected by positioning errors), Figure \ref{fig:figure1}.
They shall be connected in a sequence such that

\begin{itemize} 
\item the path goes from a \textit{PV start} to a \textit{PV end} waypoint when moving along a PV row: in this case, the distance between \textit{PV start} and \textit{PV end}  defines  how far a \textit{PV midline} shall be followed before moving to the next one; 
\item the path goes from a \textit{PV end} to a \textit{PV start} waypoint when jumping to the next PV row: in this case, the position of \textit{PV start} relative to \textit{PV end} defines the start of the new row with respect to the previous one. 
\end{itemize} 
Currently, waypoints are manually chosen, but this process might be automated starting from a georeferenced image of the whole plant. 
\begin{figure}[t]
\includegraphics[width=\columnwidth]{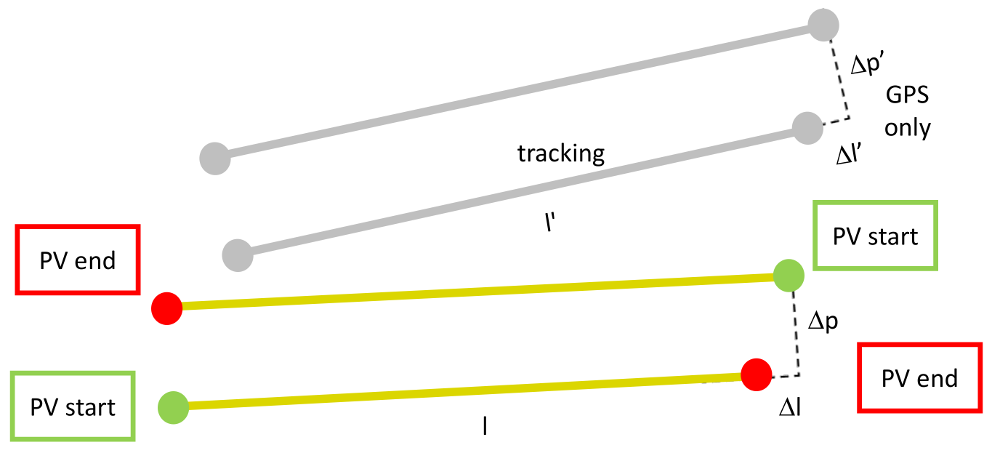}
  \caption{A priori computed waypoints (colored circles) and PV module row's position observed during the mission (gray part).}
  \label{fig:Figure3}
\end{figure}

As already discussed, real-time segmentation and visual tracking of PV module rows may play a key role in successful navigation. Indeed, waypoint-based navigation relying on the embedded positioning system of the UAV (typically merging GPS, IMUs, and compass) has well-known limitations: the absolute position of \textit{PV start} and \textit{PV end} waypoints, as well as GPS data, may be affected by large errors in the world frame. On the opposite, by using visual servoing to move along a PV row, we only need to ensure that the relative position of adjacent waypoints 
and the small UAV displacement from a PV module row to the next one performed with GPS only are sufficiently accurate. Figure \ref{fig:Figure3} shows this concept: the colored part of the Figure represents a priori computed waypoints; the gray part represents the observed position of PV module rows while tracking them. The proposed solution works even when the absolute error between the a priori computed and the observed position of PV module rows is huge, given that $\Delta l \approx \Delta l' $ and $\Delta p \approx \Delta p'$ (the additional requirement $l \approx l'$ holds if one uses GPS to measure the length of the PV module row instead of implementing some sort of ``end-of-row detector"). These assumptions look reasonable since georeferenced waypoints computed from Google Earth images and GPS data may be affected by large biases (which, in the case of GPS, vary with time) but tend to be locally coherent -- i.e., the relative error of waypoints (respectively, GPS-data) with respect to previous waypoints (respectively, GPS-data acquired in nearby locations) is small.

Eventually, even if it is not the focus of this article, remark that additional high-resolution thermal images (not used for navigation) are captured and stored in the onboard storage of the UAV for defect detection (possibly performed after the UAV has ended the mission). In this case, the UAV's flying height and speed determine the frequency of acquisition to guarantee that images are overlapping.

\section{Detection of PV modules}
\label{sec: Panel's arrays segmentation}

\subsection{Segmentation of PV modules via thermal camera}
\label{sec:Detection of PV arrays via thermal camera}

Each image captured by the thermal camera can be represented as a matrix. 
The image frame $I$  is in the upper left corner of the image plane and $i(u,v)$ is a function that associates a thermal value to each pixel $(u,v)$: thermal cameras return the thermographic image as a mono-channel grayscale intensity matrix, where the pixels intensities may assume values between 0 and 255.

\begin{figure}[t]
\includegraphics[width=\columnwidth]{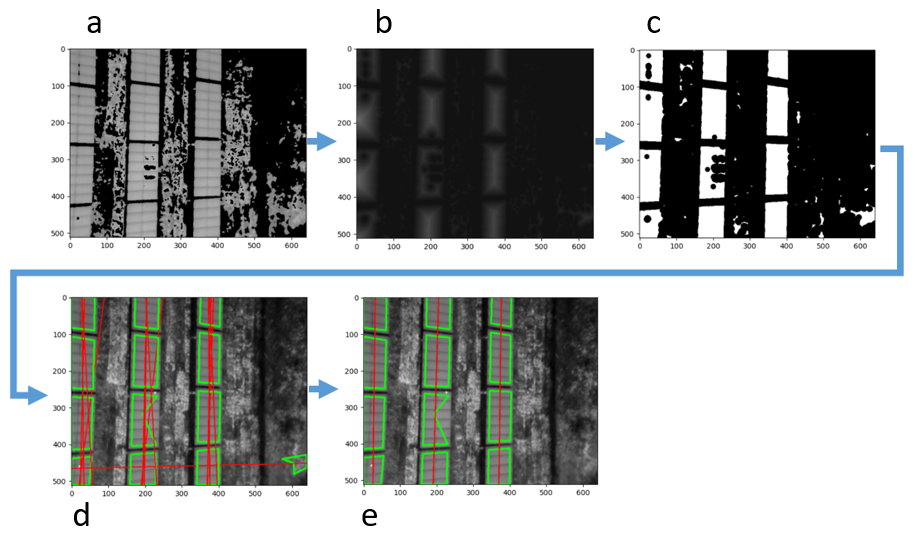}

  \caption{Pipeline for PV module detection from thermal images: a) thresholded image; b) distance matrix; c) binarized image; d) segmentation and regression line extraction; e) clustering and \textit{PV midline} estimation.}
  \label{fig:Figure4}
\end{figure}

As a first step, a mask $F$ is applied to the image after choosing two thresholds $th_1$ and $th_2$. That is, each image pixel is processed as follows:
\begin{equation}
  f(u,v) = \begin{cases} i(u,v), & \mbox{if } th_1 < i(u,v) < th_2 \\0, & \mbox{otherwise.} \end{cases}
  \label{eq1}
\end{equation}
While the values of $th_1$ and $th_2$ may be a priori chosen based on the analysis of previously acquired images\footnote{In our case, a dataset of images of a PV plant through a DJI Zenmuse H20T camera, composed of  430 $640\times 512$ thermal images.}, there may be the need to fine-tune them when environmental conditions change. Please, refer to Section \ref{sec: Detection Parameters Optimization} about the strategy used for this purpose. 
As expected and visible in Figure \ref{fig:Figure4}.a, the application of this mask is not sufficient to make the PV modules emerge unambiguously: in some regions, the ground may have the same temperature as the panels. 

As a second step, we filter out the noise by exploiting a priori domain knowledge: even if pixels with thermal intensity in the selected range can also be found outside PV panels, the panels are characterized by a higher density of pixels falling within the thresholds. 
Then, the entire image is filtered by computing a distance matrix $D$ storing the Euclidean distance of each pixel in $F$ from the closest pixel whose value is zero. 
That is, for each pixel $(u,v)$, we compute:
\begin{equation}
    d(u,v) = \sqrt{(u_{zero} - u)^2 + (v_{zero} - v)^2}
\label{eq2}
\end{equation}
where $u_{zero}$ and $v_{zero}$ are, respectively, the row  and the column of the zero value pixel nearest to $(u,v)$, Figure \ref{fig:Figure4}.b. 

As a third step, each pixel of the distance matrix $D$ is transformed into a binary intensity value as follows:
\begin{equation}
  b(u,v) = \begin{cases} 1, & \mbox{if } d(u,v) > th_3 \\ 0, & \mbox{otherwise} \end{cases}
  \label{eq3}
\end{equation}
where $th_3$ can be interpreted as a distance threshold to determine whether a pixel $(u,v)$ in the original image belongs to a high-density region, Figure \ref{fig:Figure4}.c. 
As for $th_1$ and $th_2$, the threshold $th_3$ is set a priori and then fine-tuned with the optimization procedure described in Section \ref{sec: Detection Parameters Optimization}.
Groups of neighbouring pixels are clustered into $N$ polygonal regions $S_i$: during the process, only the regions larger than a given number of pixels survive, whereas smaller clusters are deleted\footnote{OpenCv function \texttt{findcontours}.}.
The distance matrix $D$ is quite a powerful tool for estimating the position of the arrays in the image. Still, the Figure shows that it can fail, especially when the surface of the panels is not equally heated, presenting regions with different temperatures. The next step addresses this problem. 

As a fourth step, we compute, for each segmented region $S_i$, 
a regression line through polynomial curve fitting\footnote{OpenCv function \texttt{fitLine}.}: 
starting from regions $S_i$, 
$N$ regression lines are computed, each described by a point $p_i=(u_i, v_i)$ and a unitary vector $l_i$ defining a direction in the image plane $I$, Figure \ref{fig:Figure4}.d.

As a fifth step, the detection algorithm requires clustering all regions $S_i$ that are likely to correspond to PV modules aligned in the same row, by using their corresponding regression lines. 
That is, starting from a subset of individual shapes $S_i$ 
and the corresponding regression lines, a unique cluster $C_j$ (more robust to noise) is computed. 
Specifically, two regions $S_i$ and $S_k$ are clustered in $C_j$ if 
\begin{itemize}
    \item the corresponding regression lines tend to be parallel;
   \item the average point-line distance between all pixels in the image plane belonging to the first line and the second line is below a threshold, i.e., the two lines tend to be close to each other.
\end{itemize}

The clustering technique is iteratively applied for all the shapes $S_i$: then, a new regression line is computed for each cluster $C_j$, Figure \ref{fig:Figure4}.e.
The final output of the process is a set of $J$ lines, each possibly corresponding to multiple PV modules belonging to the same row. Each observed \textit{PV midline} is represented in the image frame through a couple of points $p'_j=(u'_j, v'_j)$ and $p''_j=(u''_j, v''_j)$ corresponding to the intersection of the line with the borders of the image. 




\subsection{Segmentation of PV modules via RGB camera}
\label{sec: Detection of PV arrays via RGB camera}

\begin{figure}[t] 
\centering
\includegraphics[width=\columnwidth]{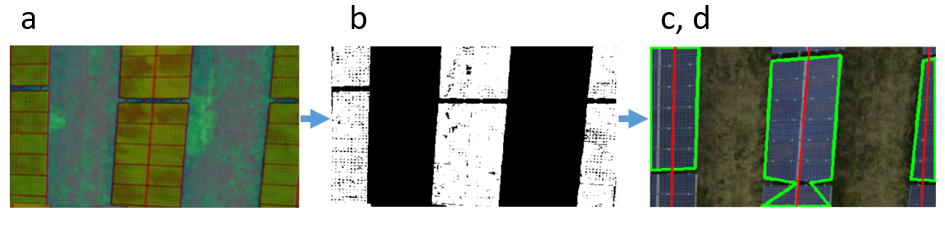}
  \caption{Pipeline for PV module detection from RGB images: a) Transformation into HSV space; b) binarized image; c,d) segmentation, regression line extraction and clustering, PV midlines estimation.}
\label{fig:Figure5}
\end{figure}

The algorithm developed to process thermal images is almost completely reused for RGB images, except for the initial steps required for PV module segmentations, Figures \ref{fig:Figure5}.a and \ref{fig:Figure5}.b.
 
When using RGB images, the rows of PV modules can be detected thanks to their color in contrast with the background. 
Then, as a first step, the original image $I$ is transformed into the hue-saturation-value (HSV) color space: Figure \ref{fig:Figure5}.a shows the result, RGB-rendered by using H values for the R channel, S for G, and V for B. 
As a second step, the PV modules are segmented from the background by thresholding the image in all the three HSV channels, Figure \ref{fig:Figure5}.b.
The lower HSV thresholds $th_4$, $th_5$, $th_6$ and the higher HSV thresholds $th_7$, $th_8$, $th_9$ are manually chosen based on a priori domain knowledge and then fine-tuned before operations, as explained in Section \ref{sec: Detection Parameters Optimization}. 

The result is a binary image defined as follows:  
\begin{equation}
  b(u,v) = \begin{cases} 1, & \mbox{if } th_4 < H(u,v) < th_7 \\
  & \mbox{$\land$ } th_5 < S(u,v) < th_8 \\ 
  & \mbox{$\land$ } th_6 < V(u,v) < th_9   \\ 0, & \mbox{otherwise} \end{cases}
  \label{eqRGB}
\end{equation}

Please remember that, in the HSV color space, hue is measured in degrees from 0 to 360 and is periodic\footnote{In OpenCV, hue is encoded in a byte and ranges from 0 to 179.}: red takes negative and positive values around zero, which means that the first condition in \eqref{eqRGB} must be changed to ``$\mbox{if not} (th_4 \leq H(u,v) \leq th_7)$" in case we need to segment areas characterized by a reddish color.

The final steps of the process, as for thermal images, require to segment polynomial regions $S_i,...,S_N$ in $b(u,v)$, each associated with a regression line, that are finally clustered into $J$ clusters $C_j$, each corresponding to an observed line $o_j=(p'_j, p''_j)$, 
Figures \ref{fig:Figure5}.c and \ref{fig:Figure5}.d.

\subsection{Thresholds tuning}
\label{sec: Detection Parameters Optimization}

PV module detection needs appropriate thresholding. For thermal images, we must choose the lower and upper bounds $th_1$ and $th_2$ of thermal intensity in \eqref{eq1} and the distance threshold $th_3$ in \eqref{eq3}; for RGB images, we need to choose the minimum $th_4$, $th_5$, $th_6$ and the maximum $th_7$, $th_8$, $th_9$ values for each HSV channel in \eqref{eqRGB}. 
Thresholds are chosen using a procedure that maximizes the number of PV modules detected in thermal and RGB images separately taken as well as the matches between the PV modules detected in both images. 
The following assumes that thermal and RGB images overlap, which can be ensured through rototranslation and scaling if needed.
Under this constraint, despite the segmented PV module shapes 
can be slightly different depending on the acquisition method used\footnote{In  thermal images, the temperature of the junctions between adjacent PV panels is usually different from PV panels themselves, resulting in a sequence of smaller rectangular shapes. In RGB images this never happens: segmentation returns larger rectangular areas composed of multiple panels.}, they are expected to roughly occupy the same regions in pairwise images.

The algorithm used to minimize the cost function is  L-BFGS-B \cite{Zhu1997550}, a limited-memory algorithm that can be used to minimize a non-linear cost function subject to simple bounds 
on the variables. The algorithm is particularly appropriate to solve large non-linear, non-convex optimization problems in which the Hessian is hard to compute. 
%
%


We assume that the UAV reaches the start position of the first PV module, possibly remotely controlled by a human operator, with its front oriented towards the direction of the PV modules row. Then, the following non-linear, non-convex optimization problem on variables $th_i$, $i=1\ldots9$, is solved: 

\begin{equation}
    \min_{th_1 - th_9} - \sum_{i = 1}^{N^T} ( c_{i}^T + r_{i}^{TR} ) - \sum_{i = 1}^{N^R} ( c_{i}^R + r_{i}^{RT} )
    \label{minRGB}
\end{equation}
where \eqref{minRGB} is the cost function, whose value expresses the ``quality" of the segmentation algorithm given a set of thresholds $th_1 - th_9$ subject to disequality constraints. 

Mathematical details of the cost are not discussed for the sake of brevity. Intuitively, please remember that the thermal camera computes a set of $N^T$ segmented regions $S_i$, $i=1\ldots N^T$, and the RGB camera computes a set of $N^R$ segmented regions $S_i$, $i=1\ldots N^R$. Then the first sum in the cost function refers to regions extracted from the thermal camera, and the second sum refers to regions extracted from the RGB camera. For each region $S_i$ extracted from thermal images, the cost function comprises two terms $c_{i}^T$ and $r_{i}^{TR}$. The term $c_{i}^T$ intuitively considers how close $S_i$ is to a rectangular shape oriented along the current direction of motion and how close $S_i$'s extension is to a value that depends on the PV module geometry and the UAV height from the ground. The term $r_{i}^{TR}$ measures the pixel-wise correlation between the region $S_i$ and the binary mask extracted from the paired RGB image covering the same area: $r_{i}^{TR}$ is higher if the region $S_i$ segmented from the thermal image is also present in the RGB image. The two terms $c_{i}^R$ and $r_{i}^{RT}$ are computed similarly, starting from regions extracted from RGB images. 

Please notice that jointly optimizing all variables might be time-consuming: for this reason, during experiments in PV plants, we decompose the problem \eqref{minRGB} into two sub-problems. Firstly, we only consider RGB images and search for the minimum of $- \sum_{i = 1}^{N^R} c_{i}^R$: that is, we ignore correlation with thermal images. Secondly, we consider thermal images only and search for the minimum of $ - \sum_{i = 1}^{N_T} ( c_{i}^T + r_{i}^{TR} )$. This simplification seems a good compromise to reduce computation time if we want to repeat the procedure, e.g., at the \textit{PV start} of each new row.


\section{UAV Navigation}
\label{sec: UAV Navigation}
\subsection{From the image frame I to the camera frame C}
\label{sec: Export of the Control Points from the Image Frame I' to the Body Frame B}
\begin{figure}[t!]
\centering
  \includegraphics[width=8cm,keepaspectratio]{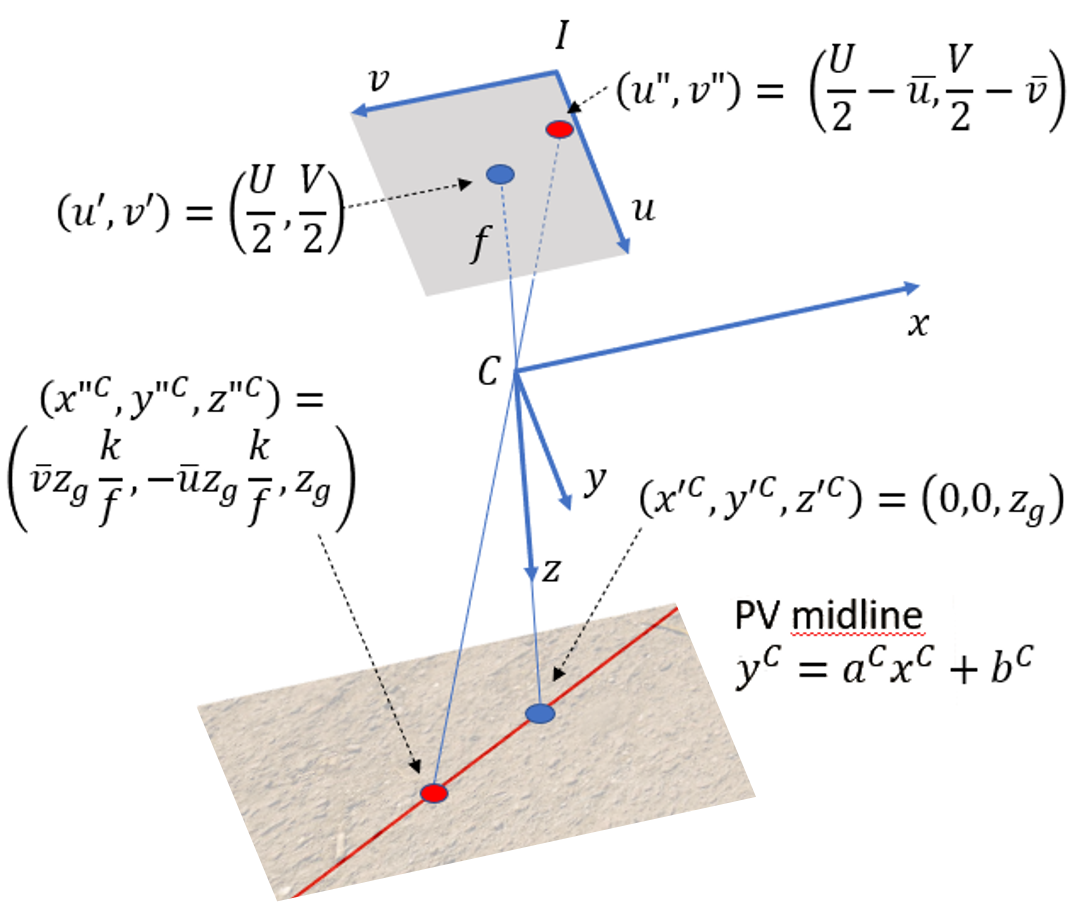}
  \caption{Converting $p_i=(u_i, v_i)$ to its position $(x^C_i, y^C_i, z^C_i)$ in the camera frame.}
  \label{fig:Figure34}
\end{figure}

To describe the motion of the UAV body in the world frame, we define a mobile camera frame $C$ with origin in the center of the camera lens, $xy$-plane parallel to the $uv$-image plane at distance $f$ (focal length), $x$-axis pointing to the UAV front, and $z$-axis downward (that is $x$ heads towards $-v$ in the image plane, and $y$ heads towards $u$, Figure \ref{fig:Figure34}). The origin of the camera frame moves as the UAV moves along the PV module row: since the UAV is endowed with a gimbal mechanism that keeps the xy-plane of the camera frame parallel to the XY-plane of the world frame $W$, all the transformations involving the image, world, or camera frame (respectively, $I$, $C$, $W$) are simpler since the camera is always looking perpendicularly to the ground in a nadiral position. 

%
%


Given a point $p_i=(u_i, v_i)$ expressed in pixel coordinates in the image frame $I$, by supposing that the point corresponds to a feature on the ground, its position $(x^C_i, y^C_i, z^C_i)$ in the camera frame $C$ can be computed as:

\begin{equation}
 x^C_i = - \frac{k}{f}(v_i - \frac{V}{2})z_{g}
\end{equation}
\label{eq: eq6}
\begin{equation}
  y^C_i = \frac{k}{f}(u_i - \frac{U}{2})z_{g}
\end{equation}
\label{eq:eq7}
\begin{equation}
  z^C_i = z_{g}
\end{equation}
\label{eq:eq7}
where $U$ and $V$ denote, respectively, the axes of the image frame,  $k$ converts the pixel size in meters, $z_{g}>0$ is the distance of the ground from the origin of the camera frame and $f<0$ is the focal length. The thermal and RGB cameras are typically mounted on the same rigid body a few centimeters apart in commercial products. Then, if we set $C$'s origin in the thermal camera lens, all observations done in the RGB image frame should ideally undergo a transformation to map them in $C$. This is not mentioned in the following for brevity's sake. 
Using the formula above, for every observed line described by $p'_i=(u'_i, v'_i)$ and $p''=(u''_i, v''_i)$ in the image frame $I$ returned by the procedure in Section \ref{sec: Panel's arrays segmentation} (either by the thermal or RGB camera), we compute the corresponding points in the camera frame $C$ and then the parameters $o_i^C=(a^C_i, b^C_i)$ of an observed line $y^C - a^C_i x^C - b^C_i=0$ in $C$ through simple geometrical considerations. Please notice that we can ignore the $z^C$ coordinate here since the $xy$ plane of the camera frame is parallel to the ground thanks to gimbal stabilization: the UAV altitude will be separately controlled through an independent mechanism. Then, the equation $y^C - a^C_i x^C - b^C_i=0$ should be better interpreted as the equation of a plane perpendicular to the $xy$ plane of the camera frame $C$.





\subsection{Path estimation through EKF}
\label{sec: Estimation of the PV array midline}
The sequence of observations $o^C_k=(a^C_k, b^C_k)$ acquired in subsequent time steps $k$, either extracted from the same image or subsequent images, are iteratively merged to estimate the actual \textit{PV midline} in the world frame. Specifically, we represent the actual \textit{PV midline} through parameters $m^W=(a^W, b^W)$, which describe a straight line in $W$ with implicit equation $y^W - a^W x^W-b^W=0$ (once again, this should be better interpreted as a plane perpendicular to the ground).
For this purpose, an Extended Kalman Filter \cite{Daum2015} is used. Please notice that the state $m^W=(a^W, b^W)$ of the \textit{PV midline} is described in the world frame $W$ to keep the system dynamics constant: thanks to this choice, the state $m^W$ to be estimated does not change with respect to $W$ as the UAV moves along the PV module row. However, the observations $o^C_k=(a^C_k, b^C_k)$ acquired through cameras at steps $k$ are expressed in the camera frame $C$; to correct the state's estimate $m^W$ through observations $o^C$, the system will need to map observations onto the world frame $W$ through the EKF observation matrix $H$, which, on its turn, requires to know the UAV pose. 

In this work, we assume that the pose $x^W, y^W$ and yaw $\theta^W$ of the camera frame in the world frame is computed by the low-level flight controller embedded in the UAV, which typically merges GPS, IMUs, and compass. Once again, thanks to the gimbal mechanism and the fact that we control the altitude through an independent mechanism, $z^W$ is ignored since the pitch and roll of the camera can be approximated to be zero. Using the embedded UAV positioning system may be counter-intuitive, since we repeatedly stated that it is affected by absolute errors. However, similar to what happens in the passage from one PV module row to the next one (Figure \ref{fig:Figure3}), it is deemed appropriate since we are not interested in knowing the absolute UAV pose with high accuracy. It is sufficient that the pose error slowly changes in time and that it is almost constant along one PV module row (which typically happens with GPS low-frequency errors) to guarantee that subsequent observations are coherent with each other when mapped to the world frame without producing abrupt changes in the state estimate\footnote{Ideally, we could attempt to estimate the UAV pose in the world frame by merging the sensors above with the observations made with cameras, by hypothesizing an augmented state vector  including both the PV midline and the UAV pose \cite{Capezio2005}. However, this possibility is not explored in this work.}.

The EKF allows for merging observations acquired at different times originating from the thermal and RGB camera.
In the following, we use the notation $\hat m^W_{k+1|k} = \hat m^W_{k|k}$ to describe the evolution of the state at the prediction step $k+1$ due to control inputs (which, as already mentioned, is constant), and the notation $o^C_{k+1} = (a^C_{k+1}, b^C_{k+1})$ to describe the observation made at the update step $k+1$ by the thermal or RGB sensor. 

During the update step, we need 
a measurement model $\hat{o}^C_{k+1} = H_{k+1}  \hat{m}^W_{k+1|k}$, where $H_{k+1}$ is the Jacobian of the non-linear observability function $h(m)$, which allows for obtaining  the expected measurement $\hat{o}^C_{k+1}$ from the estimated state $\hat{m}^W_{k+1|k}$ given the current camera position and yaw $x^{W}, y^{W}, \theta^W$. 

Specifically, $h(m^W)$ has two components: 

\begin{equation}
\begin{aligned}
     h_1(m^W) = \frac{a^W - \tan{\theta^W}}{1 + a^W \tan{\theta^W} } \\
     h_2(m^W) = \frac{x^W a^W + b^W - y^W}{\cos{\theta} + \sin{\theta}a^W}
     \label{measurementfunction}
\end{aligned}
\end{equation}
%
whose Jacobian with respect to $(a^W, b^W)$ needs to be evaluated in $\hat a^W, \hat b^W, x^W, y^W$ and $\theta^W$. 

%
%

The EKF is a recursive algorithm consisting in two steps referred to as state prediction and update. 
The prediction at step $k+1$ can be expressed as:
\begin{equation}
\hat m^W_{k+1|k} = A \hat m^W_{k|k} + B u_{k|k},
\end{equation}
where $A$, in our case, is the identity matrix and $B=0$ since the state $m^W$ is invariant over time when expressed in world coordinates. The state equations are linear, and the EKF is needed only because of the non-linearity of the observation model.
The error covariance matrix is 
\begin{equation}
  P_{k+1|k}= A P_{k+1|k} A^T + Q,
\end{equation}
where the process noise $Q\approx 0$. Since the state is not evolving, $Q$ should be null: still, we empirically add a very small contribution to the error covariance matrix to ensure that new observations will keep on contributing to the state estimate. 

Only the actual observations $o^C_{k+1}$ that are ``sufficiently close'' to the expected observations $\hat o^C_{k+1}$ (in the same spirit adopted for clustering in Section \ref{sec: Panel's arrays segmentation}) are considered during the update step. In contrast, outliers are rejected to avoid undesired corrections due to the detection of neighbouring PV rows running in parallel to the tracked one. 
Finally, in the correction step, the Kalman gain is computed starting from the covariance of the measurement noise $R$ and $P$ as usual:
\begin{equation}
    K_{k+1} = P_{k+1|k} H_{k+1}^T (H_{k+1} P_{k+1|k} H_{k+1}^T + R)^{-1}.
\end{equation}
The a posteriori estimate is updated as:
\begin{equation}
\hat{m}^W_{k+1|k+1} = \hat{m}^W_{k+1|k} + K_{k+1} (o^C_{k+1} - \hat o^C_{k+1}),
\end{equation}
\begin{equation}
 P_{k+1|k+1} = (I-K_{k+1} H_{k+1}) P_{k+1|k}.
\end{equation}
The covariance matrix $R$ relative to the thermal or RGB cameras is evaluated experimentally and takes into account both segmentation errors due to the procedure in Section \ref{sec: Panel's arrays segmentation} and the fact that, during an inspection, the GPS error may be subject to minor variations, therefore partially conflicting with the assumption of a constant GPS error in \eqref{measurementfunction}.

The EKF is initialized with $\hat m^W_{0|0}= (a^W, b^W)$ estimated at step $k=0$ by considering the two waypoints \textit{PV start} and \textit{PV end} of the initial PV module row. 

\subsection{Path following}
\label{The navigation control Algorithm}

Path following works in two different phases: (i) when the UAV is moving from \textit{PV start} to \textit{PV end} along a PV module row; (ii) when it is moving from \textit{PV end} of the current row to \textit{PV start} of the next one (Figure \ref{fig:Figure3}).

In phase (i), navigation is based on \textit{PV midline} tracking through visual servoing. After each iteration of the EKF, the estimated parameters describing the \textit{PV midline} in the world frame 
are mapped back onto the camera frame for path following, yielding $(a^C_{ref}, b^C_{ref})$.
Then, given the equation $y^C - a^C_{ref} x^C + b^C_{ref}=0$ in the camera frame, the distance error
\begin{equation}
e = - b^C_{ref}/(a^{C2}_{ref} + 1)^{1/2}
\end{equation} 
is computed, as well as the parallel and the perpendicular vectors to the path 
\begin{eqnarray}
V^C_{\parallel} = (1/a^C_{ref}, 1), & V^C_{\perp} = (-1, 1/a^C_{ref}).
\end{eqnarray}
Once the quantities above have been computed, different approaches can be used to control the distance from the \textit{PV midline}. In this work, we adopt a simple ``carrot chasing'' approach \cite{Sgorbissa2019}: the parallel and perpendicular vectors are added to compute the position of a virtual moving target using the distance error $e$ as a weighting factor for $V^C_{\perp}$, and a PID controller is used to tune the UAV velocities to regulate the distance from the target to zero.

In phase (ii), the UAV uses the pose information $x^W$, $y^W$, $\theta^W$ returned by the onboard flight controller (merging GPS, IMU, and compass) to move from \textit{PV end} to the next \textit{PV start}. 

In both phases, the altitude is separately controlled using embedded sensors returning the height from the ground\footnote{A method to automatically control the UAV altitude depending on the difference between the actual and expected dimensions of PV modules on the image plane was implemented, but it is not discussed in this work.}.

\section{Experiments}
\label{Experiments}
This Section presents the experiments conducted in a real PV plant and a realistic simulation environment.

\subsection{Material and Methods}
\begin{figure*} 
\centering
  \includegraphics[width=1.7\columnwidth,  keepaspectratio]{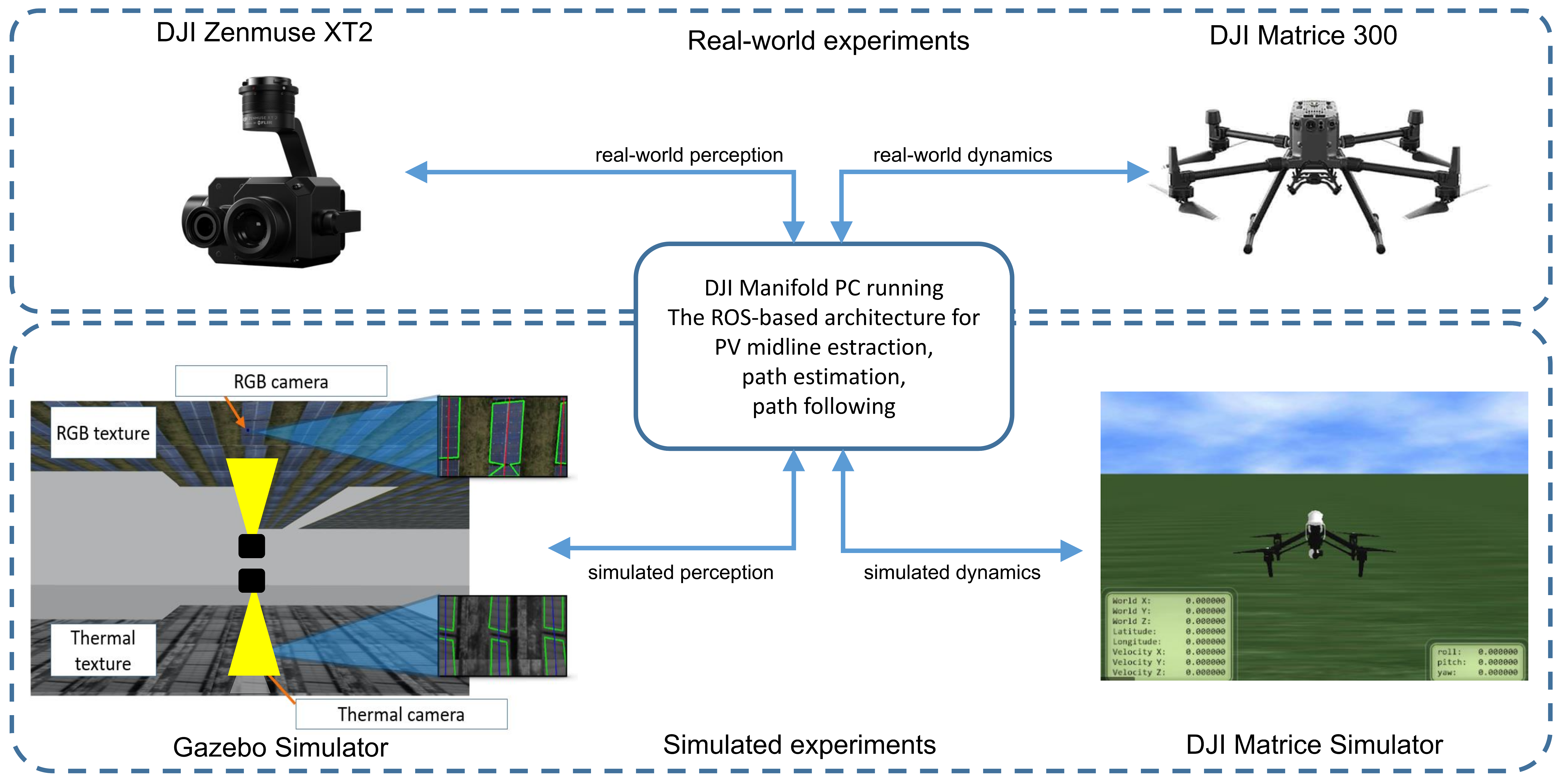}
  \caption{During real-world experiments, images are acquired through a DJI Zenmuse XT2 camera and used to control a DJI Matrice 300. During simulated experiments, images are acquired in a simulated PV plant in  Gazebo and used to control the UAV's dynamics in the 300 DJI simulator. The positions of the RGB (looking upward) and thermal (looking downward) cameras are updated accordingly to the UAV's simulated dynamics.}
  \label{fig:Figure6}

\end{figure*}

Real-world experiments in a PV plant were conducted using a DJI Matrice 300 aircraft equipped with a DJI Zenmuse XT2 camera, Figure \ref{fig:Figure6}. 
The camera has a field of view of $57.12 \times 42.44$ degrees and a gimbal mechanism. 
An onboard DJI Manifold PC equipped with an NVIDIA Jetson TX2 and 128GB of internal memory performs all the computations described in previous Sections to process the acquired images and control the UAV.

The simulation environment deserves more attention, Figure \ref{fig:Figure6}. In addition to the onboard DJI Manifold performing the core computations, 
two programs were running in parallel to simulate, respectively, the dynamics and the perception of the UAV:
\begin{itemize}
    \item \textit{The DJI Matrice Simulator embedded in the DJI Matrice 300.} When the DJI Manifold is connected to the drone and the OSDK is enabled, this program simulates the UAV dynamics based on the commands received from the ROS nodes executed on the DJI Manifold.
    \item \textit{The Gazebo Simulator, running on an external Dell XPS notebook with an Intel i7 processor and 16 GB of RAM, integrated with ROS.} Here, only the thermal and RGB camera and the related gimbal mechanism are simulated to provide the DJI Manifold with images acquired in the simulated PV plant. 
\end{itemize}

    The cameras' position inside Gazebo is linked to the UAV position returned by the DJI Matrice Simulator to guarantee coherence in the simulated images returned during the simulated flight. 
%
%
%
Specifically, RGB textures are placed ``above in the sky," while thermal textures are placed ``below on the ground." By adding two cameras in Gazebo, one directed towards the sky and the other directed towards the ground, we can simulate the acquisition of RGB and thermal images of the same PV plant. Textures were produced from images captured by the DJI Zenmuse XT2 during a PV plant inspection in northern Italy.

Both in real-world experiments and in simulation, we initially compute the ``reference  waypoint path" to be followed by the UAV by considering PV module rows one after the other in boustrophedon mode. 
Once the UAV has reached the \textit{PV start} of the first PV row, we execute the optimization process for tuning thresholds depending on environmental conditions by letting the UAV hover at a certain height over the PV module. Then the UAV starts moving using both cameras and the EKF for iteratively extracting the PV midline, correcting the reference path, and moving along it. 
When the UAV reaches \textit{PV end}, the visual tracking stops, and the UAV moves to the next \textit{PV start}. 
We consider the mission completed when the UAV has inspected the required number of parallel PV module rows. 

The thermal and RGB images are processed, at an approximate frequency of 3 and 5 fps, respectively, both in the simulated and real experiment (computations can likely be made more efficient to achieve a higher frame rate). Observations are used to correct the estimate of the \textit{PV midline} through the EKF. Then, velocity commands for path following are computed and sent to the onboard controller of the DJI Matrice 300 approximately with a 5Hz frequency in simulation (due to heavy computations performed in Gazebo) and a 30Hz frequency in real cases. 


Simulated experiments (Section \ref{sec: Simulation Experimental Results}) were mainly aimed to validate the proposed approach and assess the system's reliability in the presence of inaccurate waypoint positioning. 
Experiments in the real scenario (Section \ref{Real Experimental Results}) had the primary purpose of comparing the proposed approach's robustness and assessing advantages deriving from combining thermal and RGB cameras.





\subsection{Threshold optimization time} 
\label{sec: Optimization Experimental Results}

We analyzed the time required to perform the optimization procedure on RGB and thermal images collected during different flights with different daylight conditions. Table \ref{table:table1} shows, for five subsequent tests, the time$t_{HSV}$  required for HSV optimization (thresholds $th_4 - th_9$), the time $t_{thermal}$ required for thermal optimization (thresholds $th_1 - th_3$), and the overall time $t_{tot}$ required by the algorithm to complete the process.  

As visible from the results, the optimization algorithm is reasonably fast in finding a suitable range of threshold values to detect and cluster PV module rows in the images, also in the presence of possible misalignments between the two images. Tests show that the optimization time can vary depending on the number of PV module rows in each image and daylight conditions. However, please observe that appropriate light conditions also play a crucial role in defect detection (not addressed in this article). Flying during cloudy days or at night does not deserve to be explored because the detection of defects in the PV modules can be unreliable due to the low temperature reached by the panel surface. 

Finally, please consider that the experiments described in the following sessions were performed with images collected at different hours of the day and on different days. In all experiments, we noticed that the choice of the thresholds $th_1 - th_9$, once performed, tends to work well for the whole experiment duration, making it possible to execute the optimization algorithm only at the beginning of each flight. 

\begin{table}[t!]
\centering

 \begin{tabular}{||c c c c||} 
 \hline
 Test $\#$ & $t_{HSV}$[s] & $t_{thermal}$[s]  &  $t_{tot}$[s] \\ [0.5ex] 
 \hline\hline
 1 & 68.6 & 38.9  & 107.6 \\ 
 2 & 84.6 & 51.0 & 135.6 \\
 3 & 84.4 & 49.3 & 133.7 \\
 4 & 85.1 & 55.2 & 140.3 \\
 5 & 71.4 & 52.0 & 123.4 \\ [1ex] 
 \hline
 \end{tabular}
  \caption{Threshold selection times in different tests.}
\end{table}
 
 \label{table:table1}

\begin{figure}[t!]
\centering
\includegraphics[width=9cm, keepaspectratio]{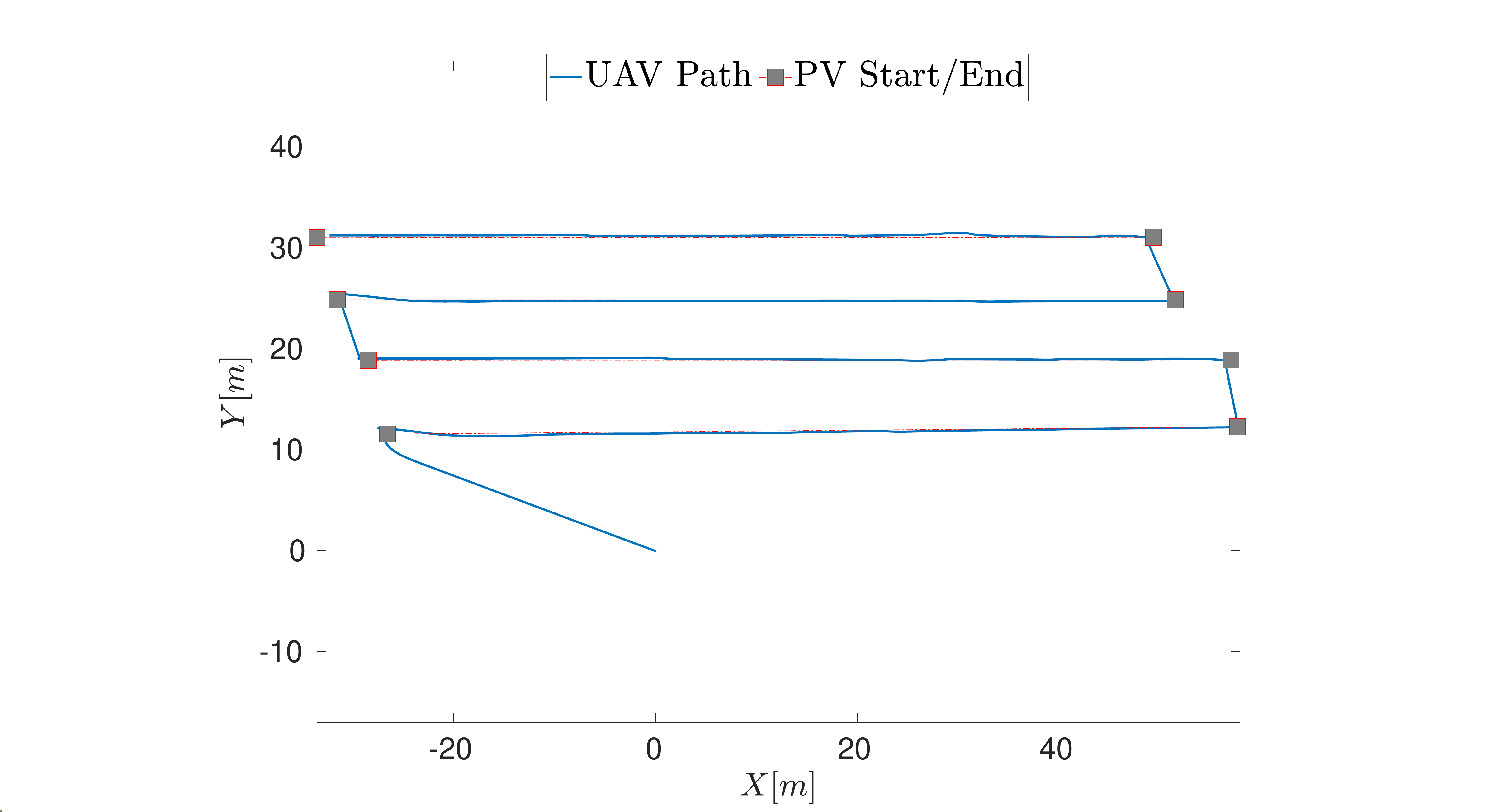}
	\caption{Projection on the XY-plane of the UAV  path (thermal camera).}
	\label{fig:Figure8}
\end{figure}

\begin{figure}[t!]
\includegraphics[width=9.5 cm,keepaspectratio]{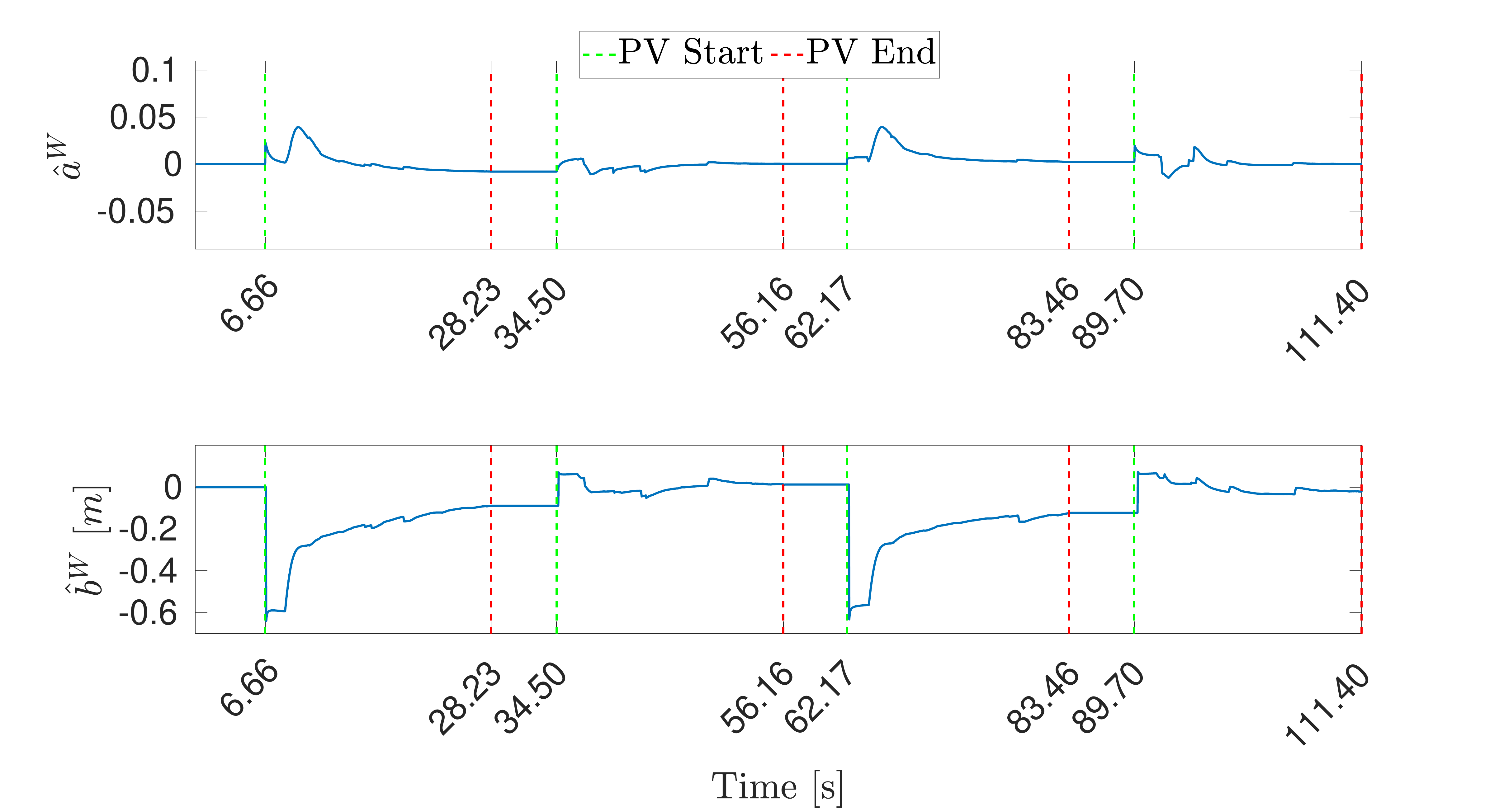}

	\caption{EKF estimation error of the \textit{PV midline} parameters $\hat a^W$ and $\hat b^W$ (thermal camera).}
	\label{fig:Figure9}
\end{figure}

\begin{figure}[t!]
\includegraphics[width=9.5 cm, keepaspectratio]{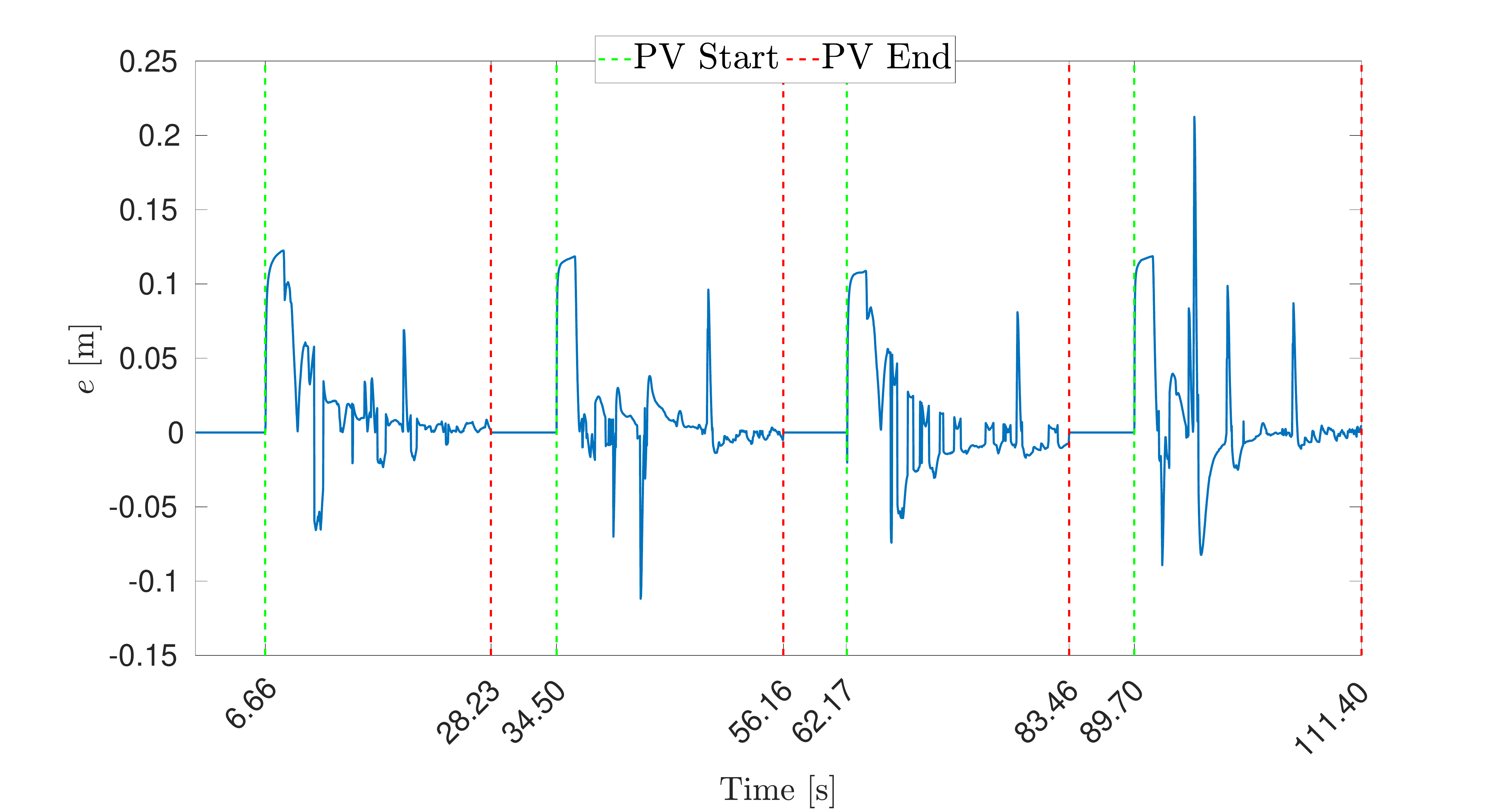}
	\caption{Control error $e$ from the estimated \textit{PV midline} (thermal camera).}
	\label{fig:Figure10}
\end{figure}

\begin{figure}[t!]
\includegraphics[width=9.5 cm, keepaspectratio]{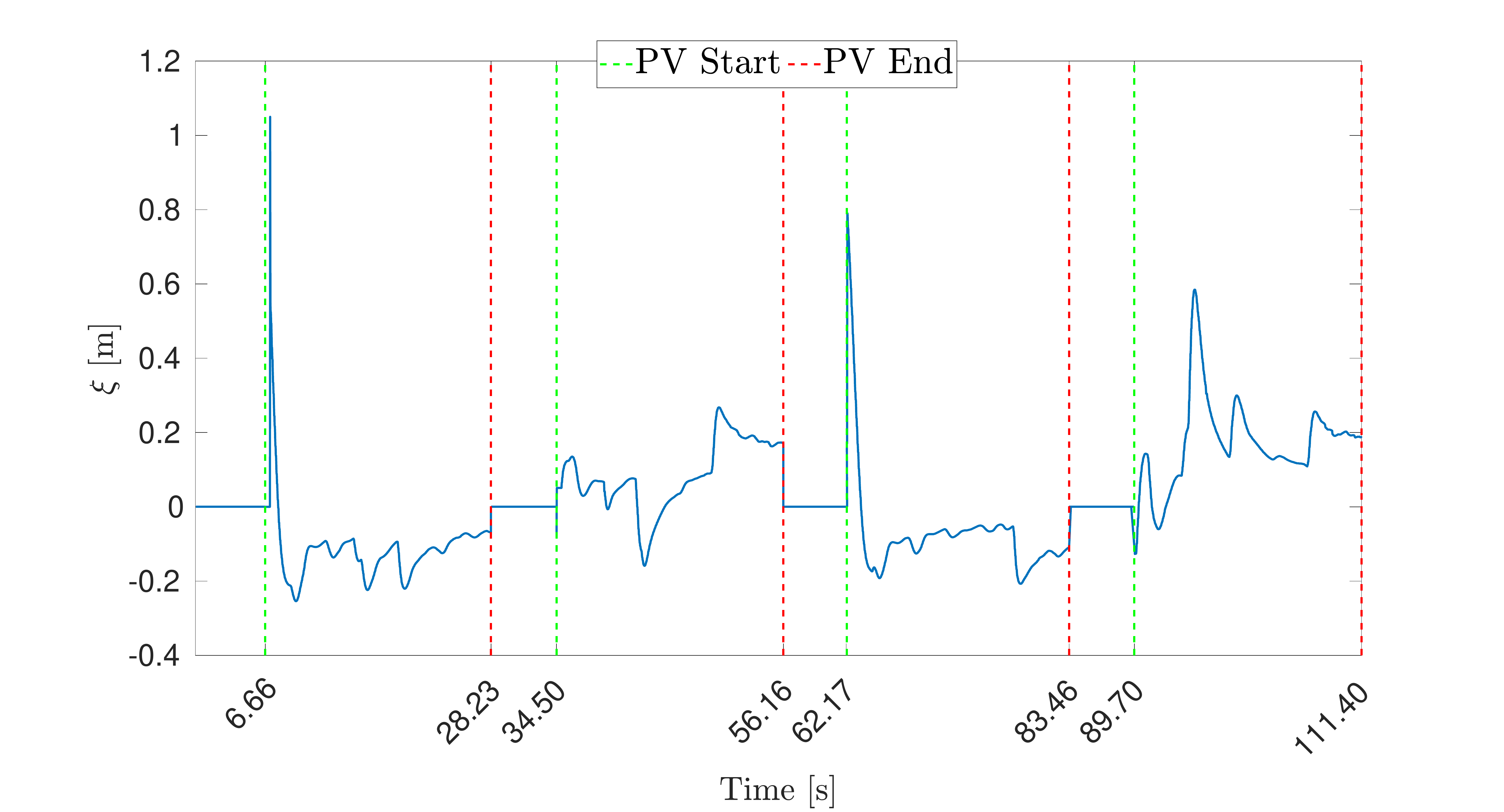}
	\caption{Navigation error $\xi$ from the real \textit{PV midline} (thermal camera).}
	\label{fig:Figure11}
\end{figure}


\subsection{Experimental Results in Simulation}
\label{sec: Simulation Experimental Results}

Two classes of experiments were performed in simulation:
\begin{itemize}
	\item  Subsection \ref{Navigation in normal light condition} reports the simulated experiments with the thermal camera only, RGB camera only, and both cameras for PV module detection. Here we do not consider errors in waypoints, which are correctly located on the midlines of the corresponding PV rows.
	
	\item Subsection \ref{Navigation with Roto-Translation error in Waypoint's position} reports the simulated experiments with both cameras to assess the robustness of the approach in the presence of errors in waypoint positioning. 
	
	
\end{itemize}


\subsubsection{Testing the impact of different cameras for panel detection}
\label{Navigation in normal light condition}

The mission consisted in inspecting four parallel PV rows. The waypoints were assumed to be correctly placed at each row's start and end without errors. 
During navigation, the UAV velocity had a constant value of $0.6$ m/s, and the UAV flew at a constant height of $15$ m from the take-off point. Please notice that, at a $15$ m height, a speed lower than $2$ m/s is appropriate to prevent blurred images and ensure image overlapping. 








Experiments were initially performed with the thermal camera only. In Figure \ref{fig:Figure8}, the projection on the XY-plane of the path followed by the UAV along four parallel PV rows (blue line) and the reference path determined by the waypoints (red dashed line) are shown. After take-off, the UAV autonomously reached the first \textit{PV start}, and hovered there for some second before moving along the panels, collecting observations to estimate the \textit{PV midline}. Navigation performance can be evaluated by measuring the convergence of the UAV position  to the midline of PV module rows.

In Figure \ref{fig:Figure9}, the error between the real \textit{PV midline} described by ${m} = (a^W, b^W)$ (the ground truth is known in simulation) and the estimated state $\hat{m} = (\hat{a}^W, \hat{b}^W)$ returned by the EKF is shown. Green vertical lines correspond to \textit{PV start}; red vertical lines correspond to \textit{PV end} of the same row. Since the UAV moved from the end of one row to the start of the next one by using positioning information provided by the onboard flight controller, the plots included between a red vertical line and the subsequent green line shall be ignored since they correspond to a change of PV module row.



%

In Figure \ref{fig:Figure10}, the control error $e$ between the UAV position and the \textit{PV midline} estimated with the EKF is shown, with average $\mu_e= 0.022$ m and standard deviation $\sigma_e=0.031$ m (we consider only values included between a green vertical line and the subsequent red vertical line).

In Figure \ref{fig:Figure11}, the navigation error $\xi$ between the UAV position and the actual \textit{PV midline} (the ground truth is known in simulation) is shown. The evaluated RMSE (Root Mean Squared Error)  during navigation, computed from $\xi$, is $0.178 $ m.
The figure shows that the navigation error tends to be larger when the UAV starts to track the PV module row after reaching \textit{PV start}, green vertical line. This is coherent with the fact that \textit{PV start} is considered to be reached when the UAV position is within a 1-meter distance, thus producing an initial error that is recovered later.  

\begin{figure}[t]
\includegraphics[width=9.5 cm, keepaspectratio]{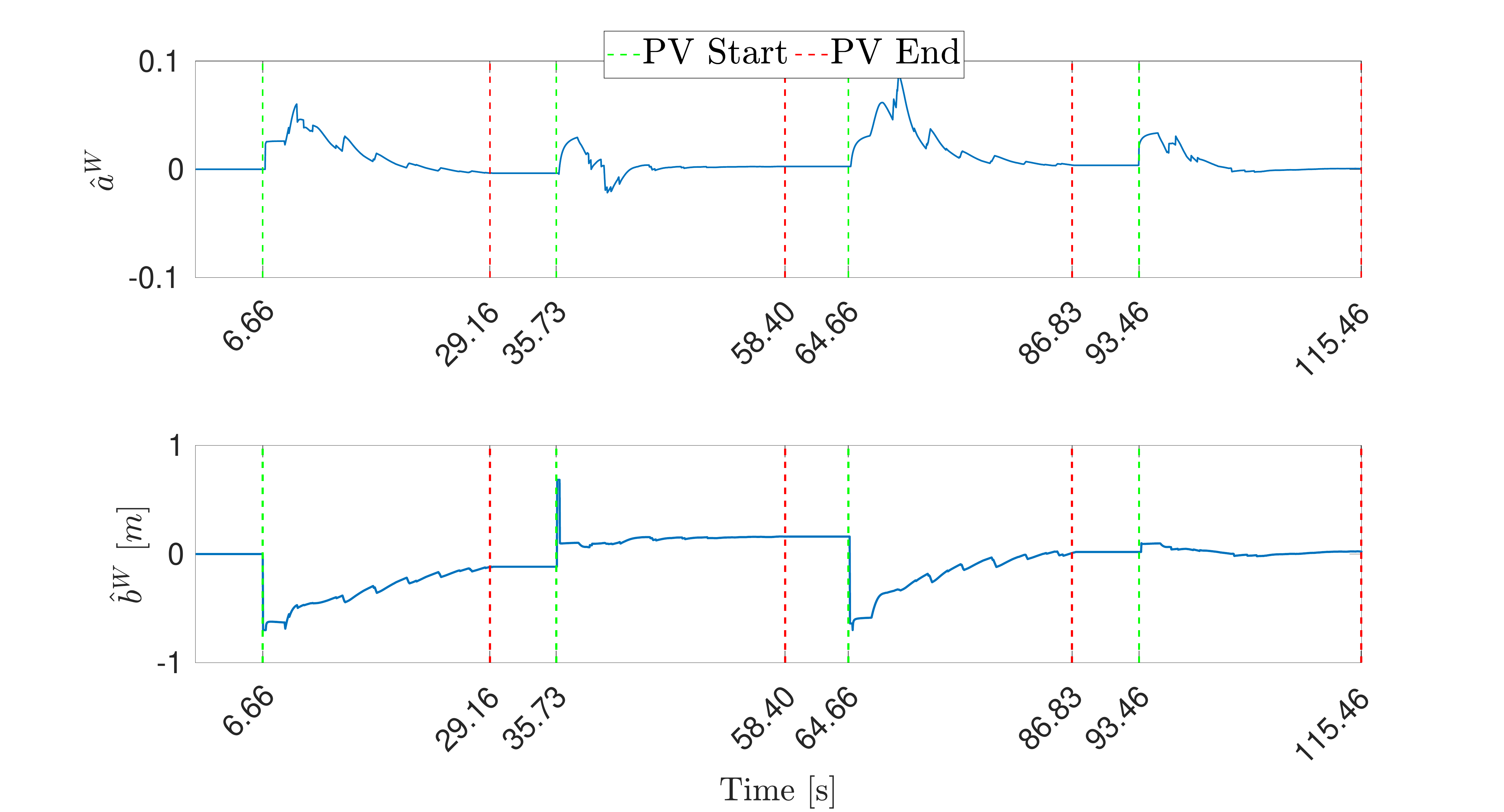}
	\caption{EKF estimation error of the \textit{PV midline} parameters $\hat a^W$ and  $\hat b^W$ (RGB camera).}
	\label{fig:Figure12}
\end{figure}

\begin{figure}[t]
\includegraphics[width=9.5 cm, keepaspectratio]{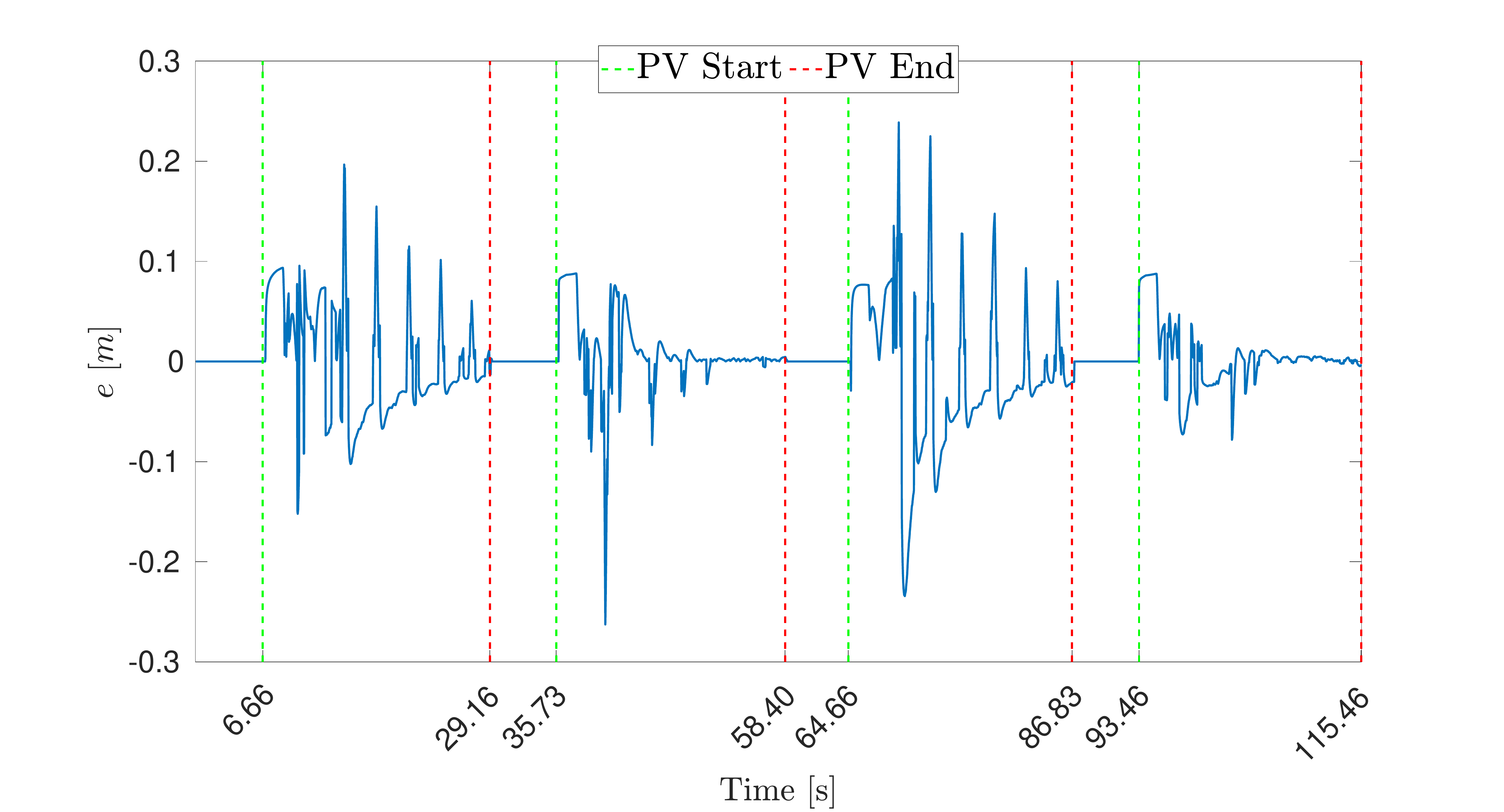}
	\caption{Control error $e$ from the estimated \textit{PV midline} (RGB camera).}
	\label{fig:Figure13}
\end{figure}

\begin{figure}[t]
\centering
\includegraphics[width=9.5 cm, keepaspectratio]{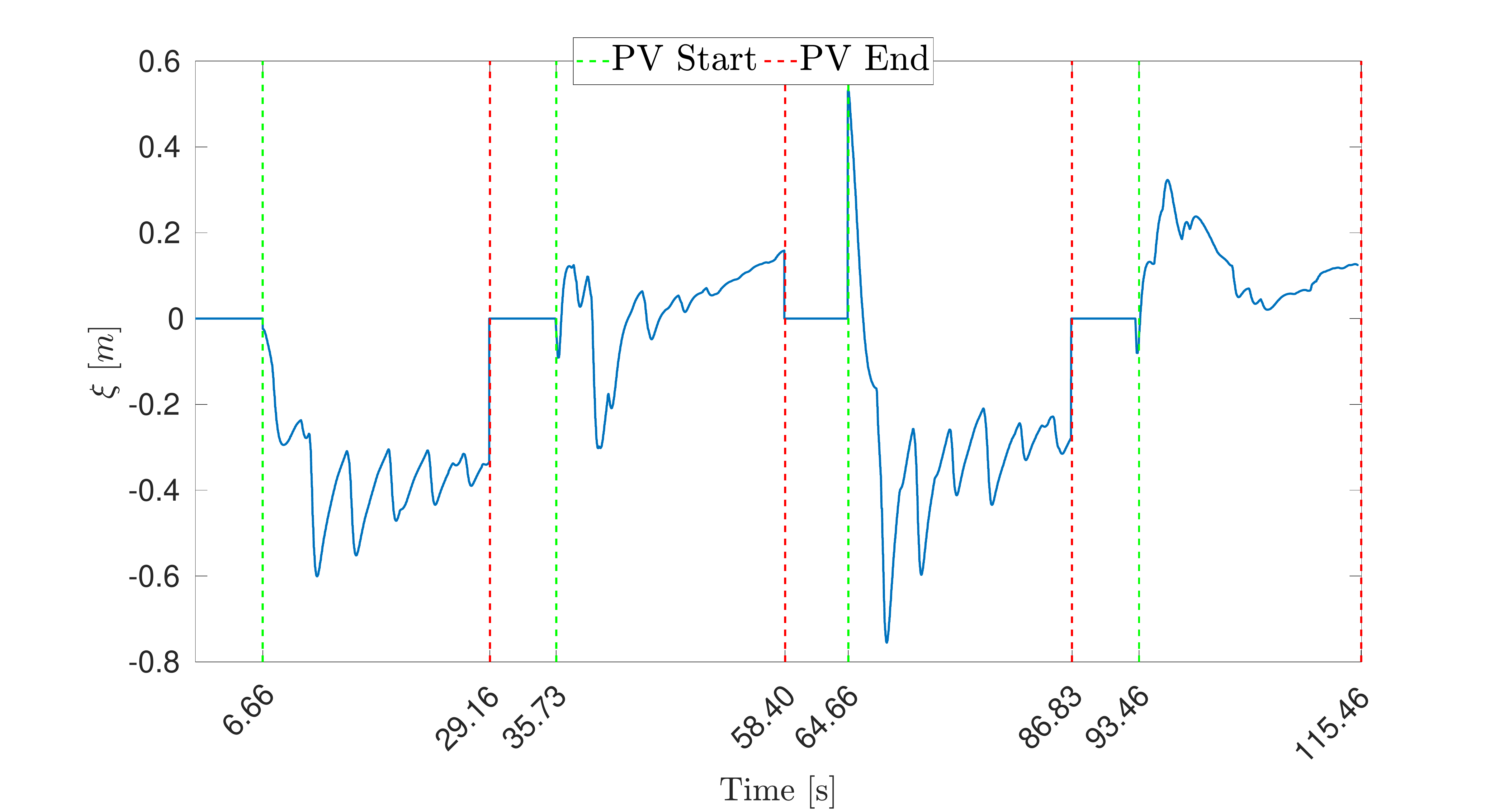}
	\caption{Navigation error $\xi$ from the real \textit{PV midline} (RGB camera).}
	\label{fig:Figure14}
\end{figure}

Next, experiments were performed with the RGB camera only. 
The projection on the XY-plane of the UAV path using the RGB camera is not reported since it is difficult to appreciate differences from the previous experiment. Figure \ref{fig:Figure12} reports the error between the actual and the estimated state of the \textit{PV  midline}. Figure \ref{fig:Figure13} reports the control error, with average $\mu_e= 0.032$ m and standard deviation $\sigma_e=0.036$ m (i.e., performance is worse than those obtained during the experiment with the thermal camera). Figure \ref{fig:Figure14} reports the navigation error, with an RMSE of $0.246$ m.

\begin{figure}
\includegraphics[width=9.5 cm, keepaspectratio]{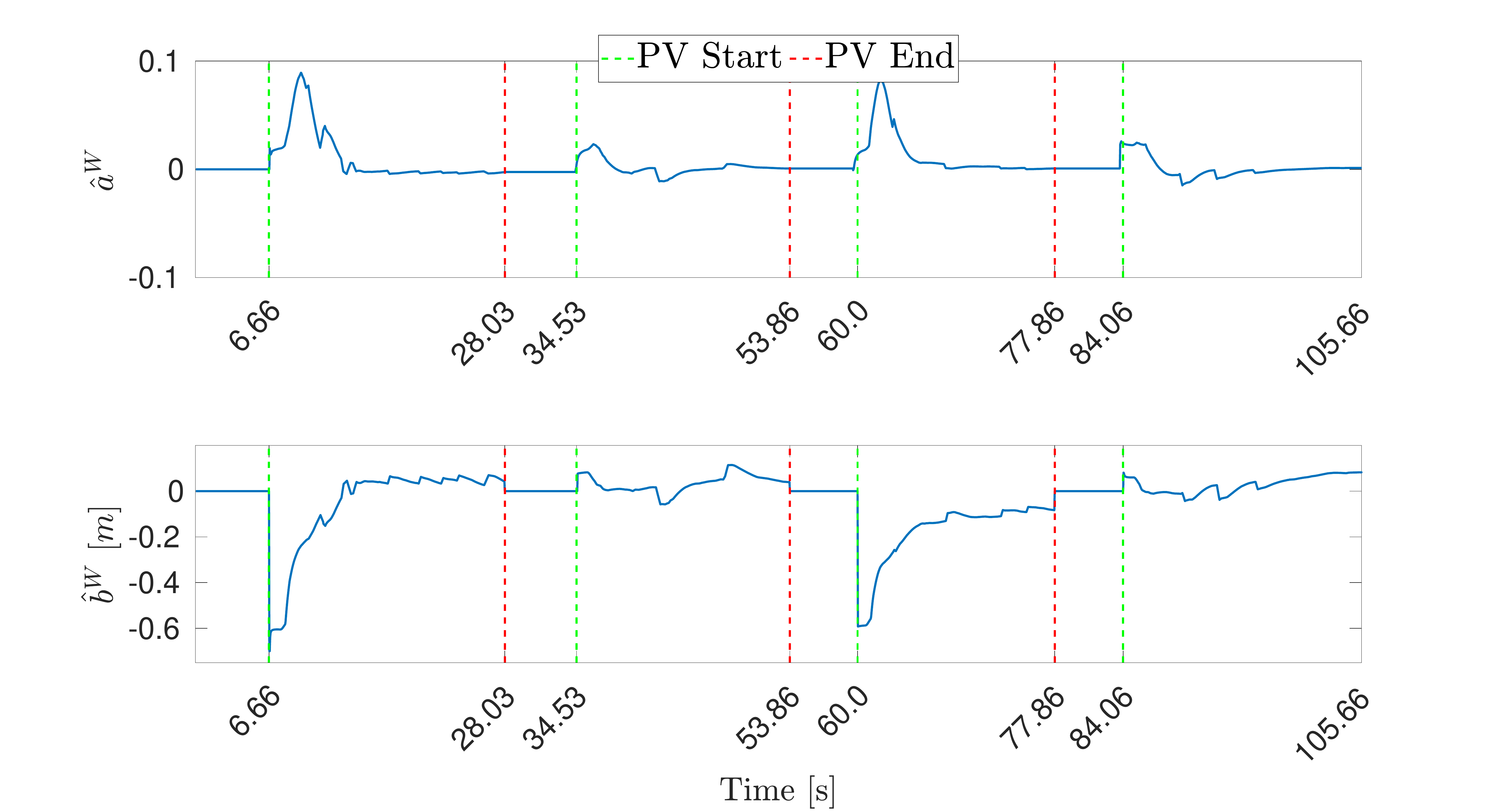}
	\caption{EKF estimation error of the \textit{PV midline} parameters $\hat a^W$ and  $\hat b^W$ (both cameras).}
	\label{fig:Figure15}
\end{figure}

\begin{figure}[t]\hfill
\includegraphics[width=9.5 cm, keepaspectratio]{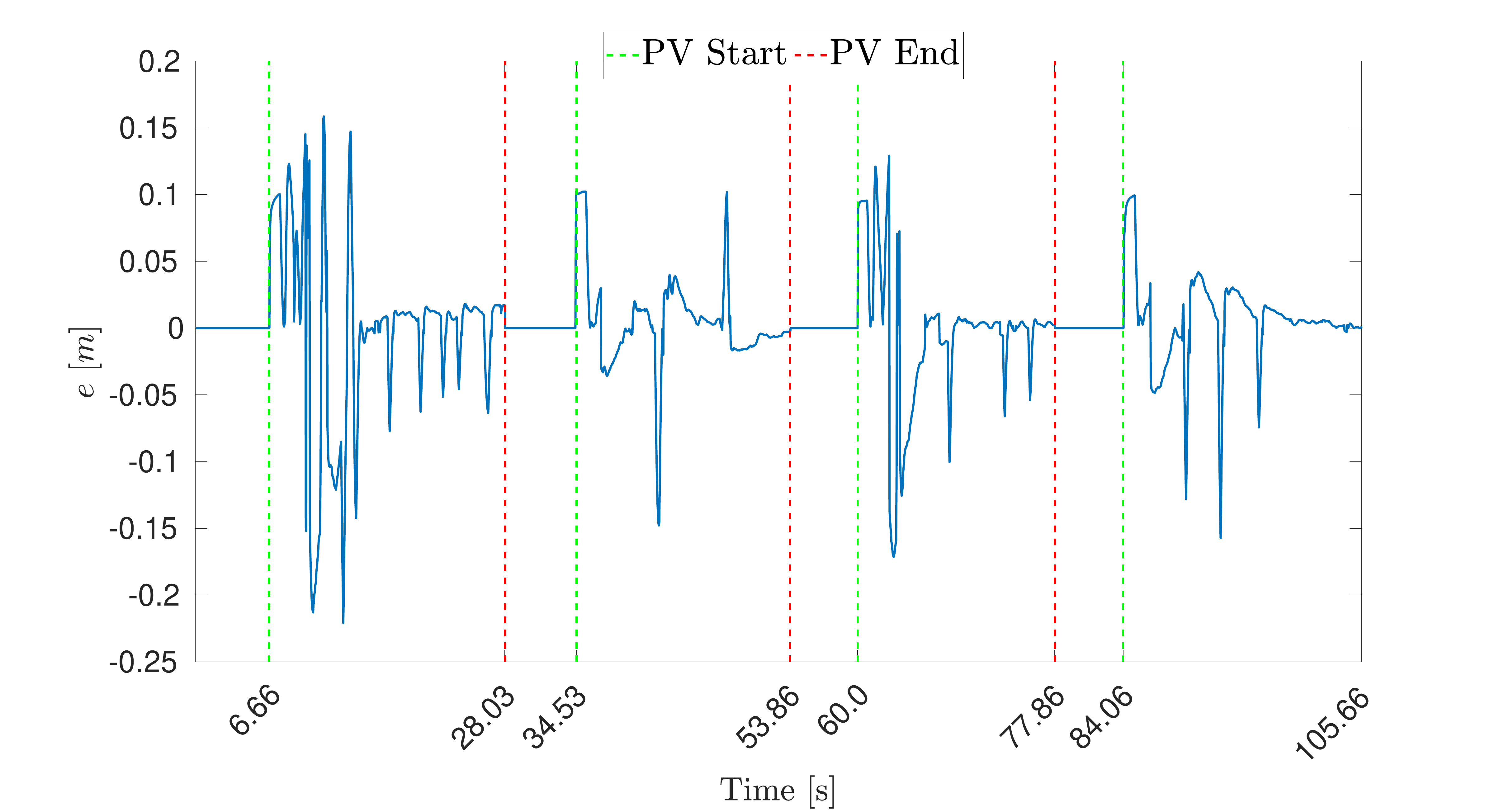}
	\caption{Control error $e$ from the estimated \textit{PV midline} (both cameras).}
	\label{fig:Figure16}
\end{figure}

\begin{figure}[t]\hfill
\includegraphics[width=9.5 cm, keepaspectratio]{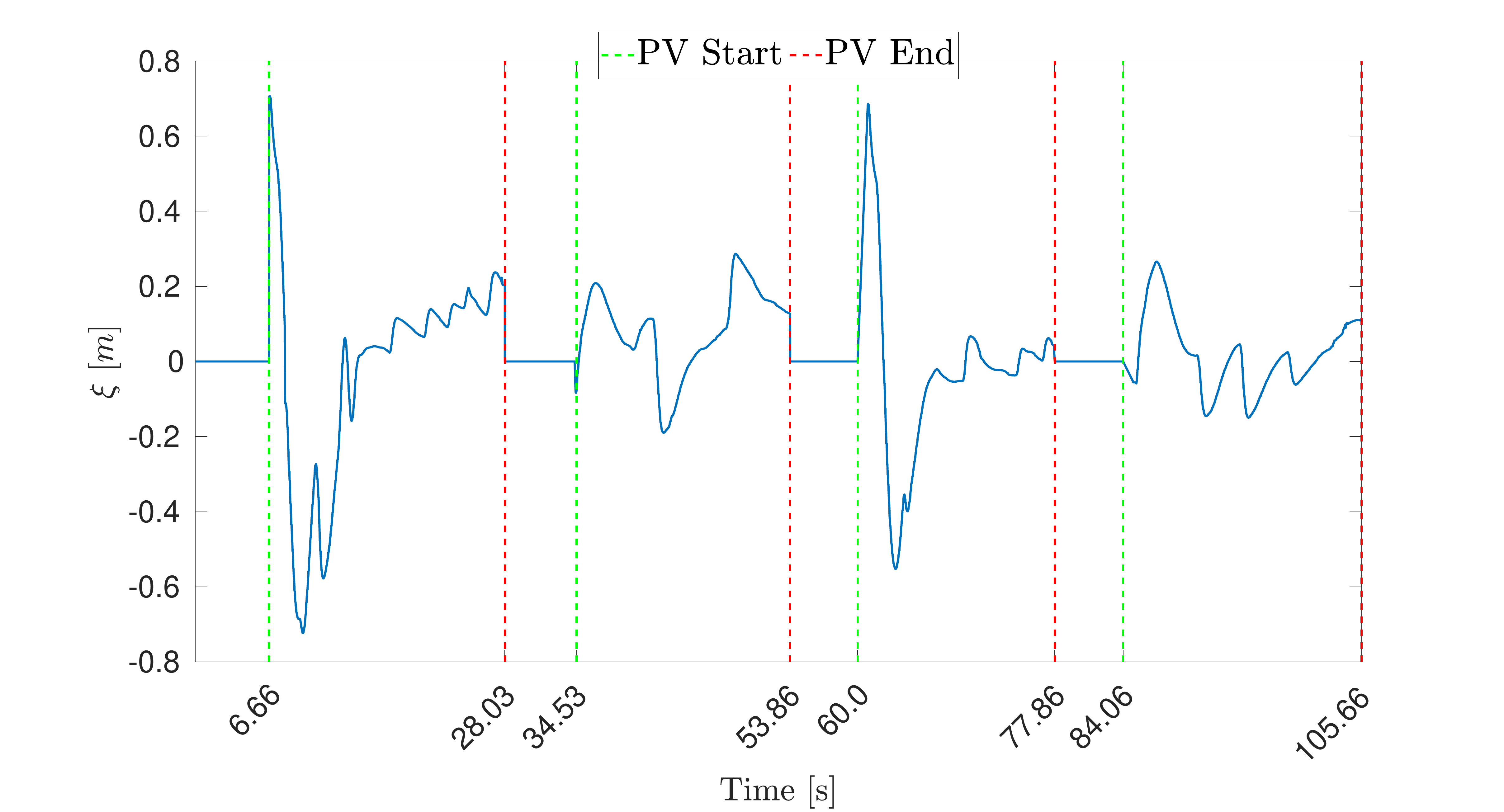}
	\caption{Navigation error $\xi$ from the real \textit{PV midline} (both cameras).}
	\label{fig:Figure17}
\end{figure}

Eventually, experiments were performed with both cameras. 
Figure \ref{fig:Figure15} reports the error between the actual and the estimated state of the \textit{PV midline}. Figure \ref{fig:Figure16} reports the control error, with average $\mu_e= 0.026$ m and standard deviation $\sigma_e=0.036$ m. Figure \ref{fig:Figure17} reports the navigation error with an RMSE of $0.153 $ m. 
As expected, the presence of both thermal and RGB cameras slightly improves navigation performance. 


%

\subsubsection{Navigation with errors in waypoint positions}
\label{Navigation with Roto-Translation error in Waypoint's position}

\begin{figure}[!t]
\includegraphics[width=9.5 cm, keepaspectratio]{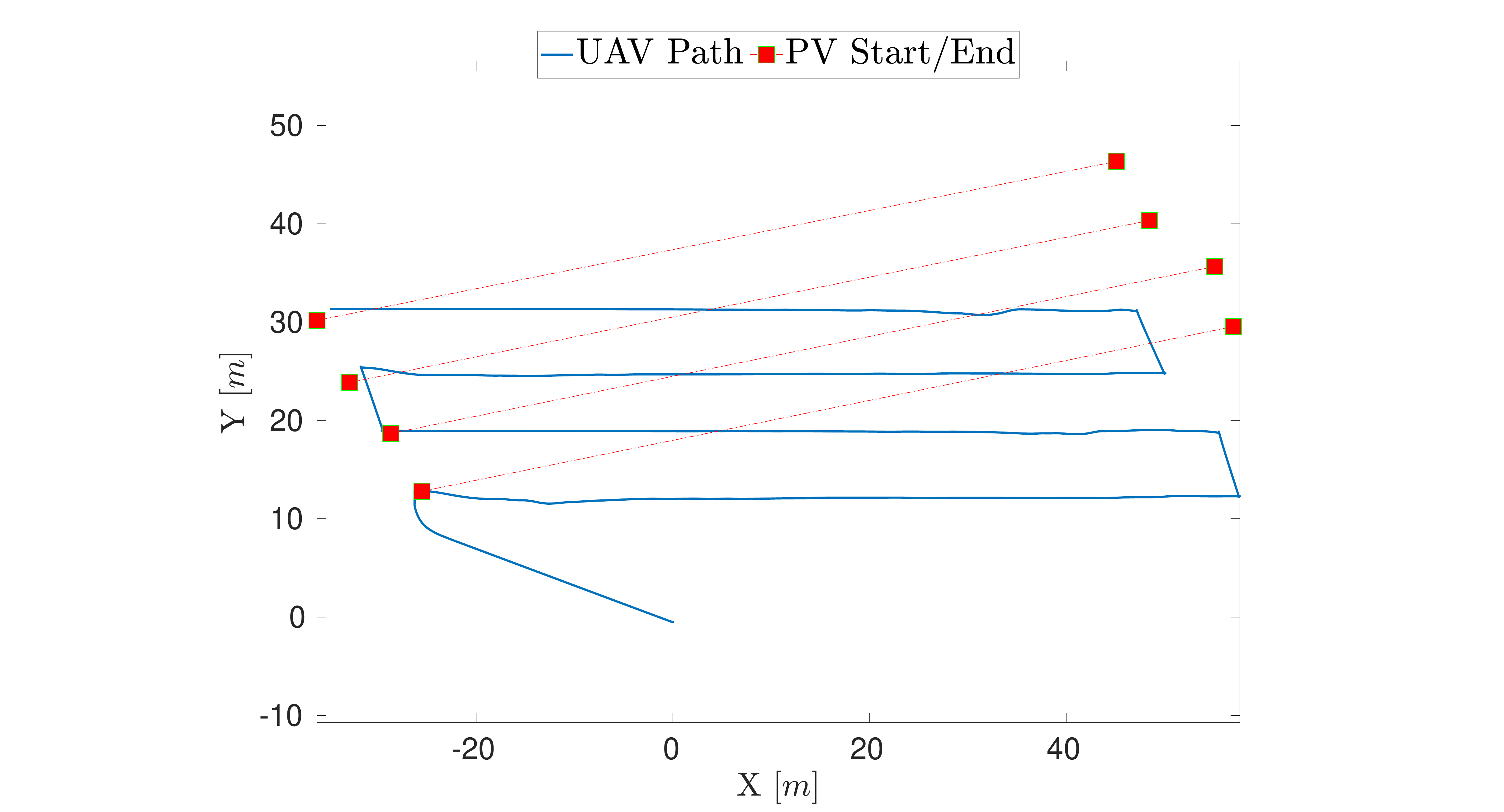}
\centering
	\caption{Projection on the XY-plane of the UAV  path in the presence of waypoint positioning errors (both cameras).}
	\label{fig:Figure18}
\end{figure}

\begin{figure}[!t]
\includegraphics[width=9.5 cm, keepaspectratio]{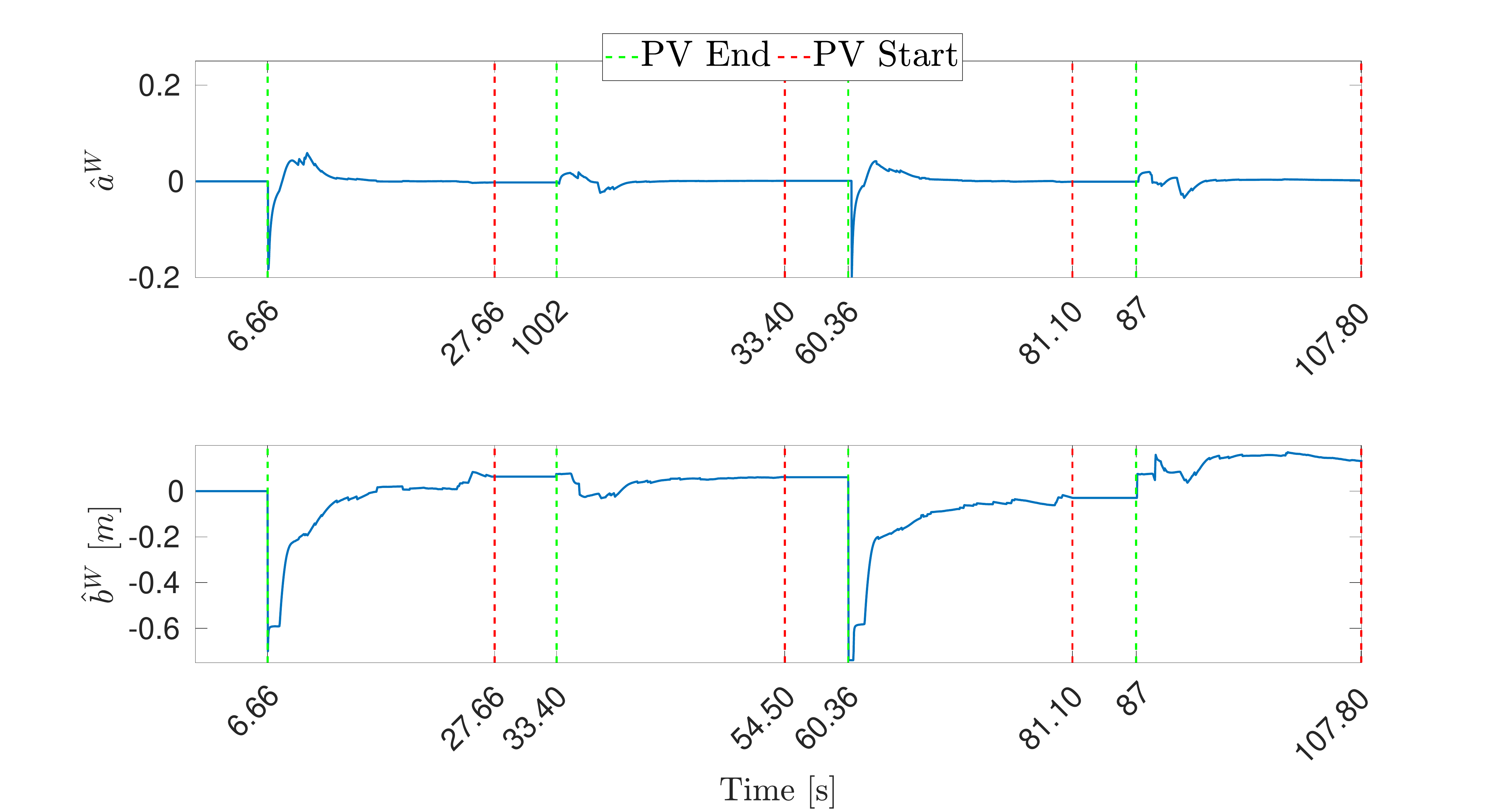}
	\caption{EKF estimation error of the \textit{PV midline} parameters $\hat a^W$ and  $\hat b^W$  in the presence of waypoint positioning errors (both cameras).}
	\label{fig:Figure19}
\end{figure}

\begin{figure}[!t]
\includegraphics[width=9.5 cm, keepaspectratio]{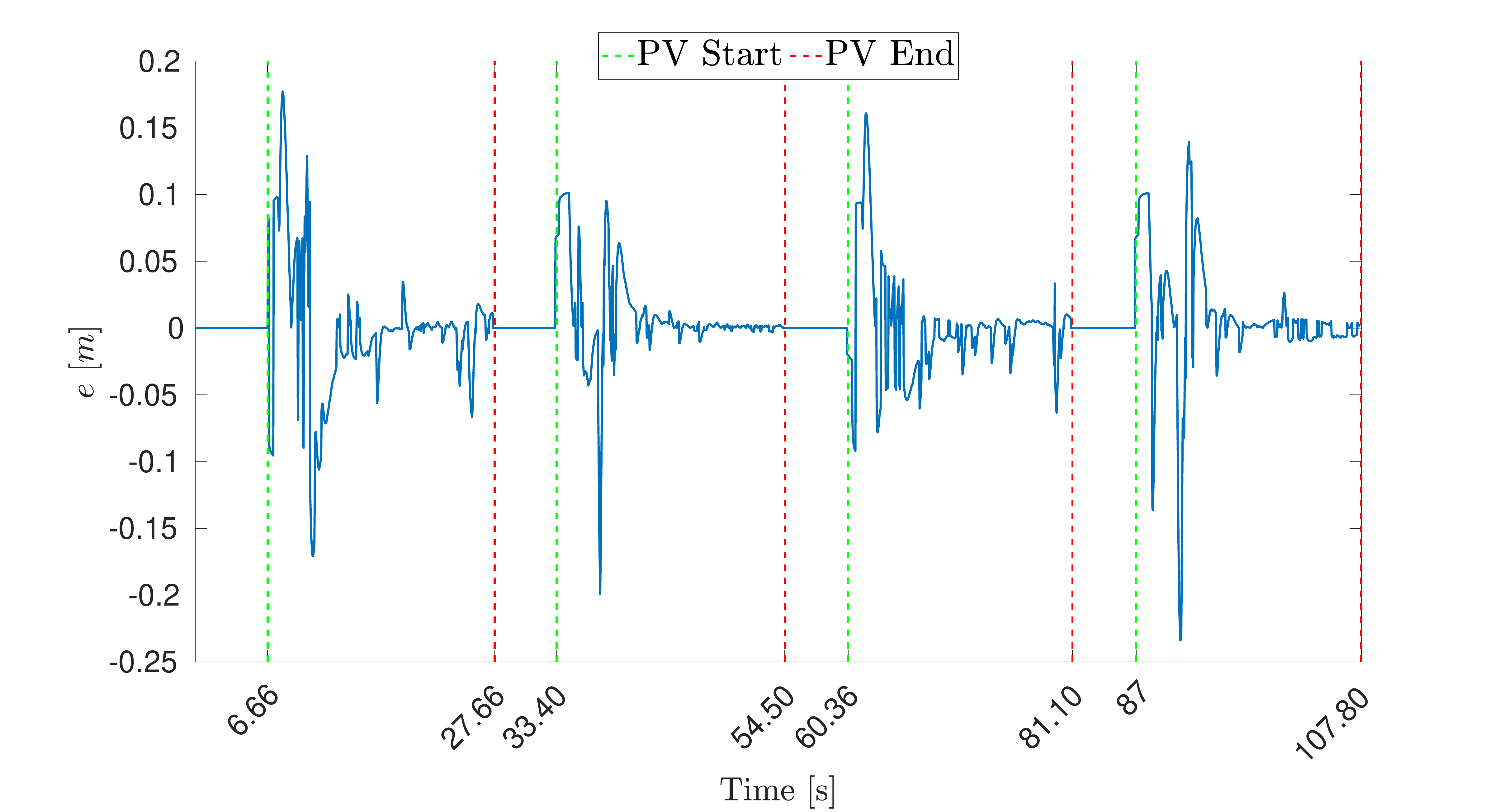}
	\caption{Control error $e$ from the estimated \textit{PV midline} in the presence of waypoint positioning errors (both cameras).}
	\label{fig:Figure20}
\end{figure}

\begin{figure}[!t]\hfill
\includegraphics[width=9.5 cm, keepaspectratio]{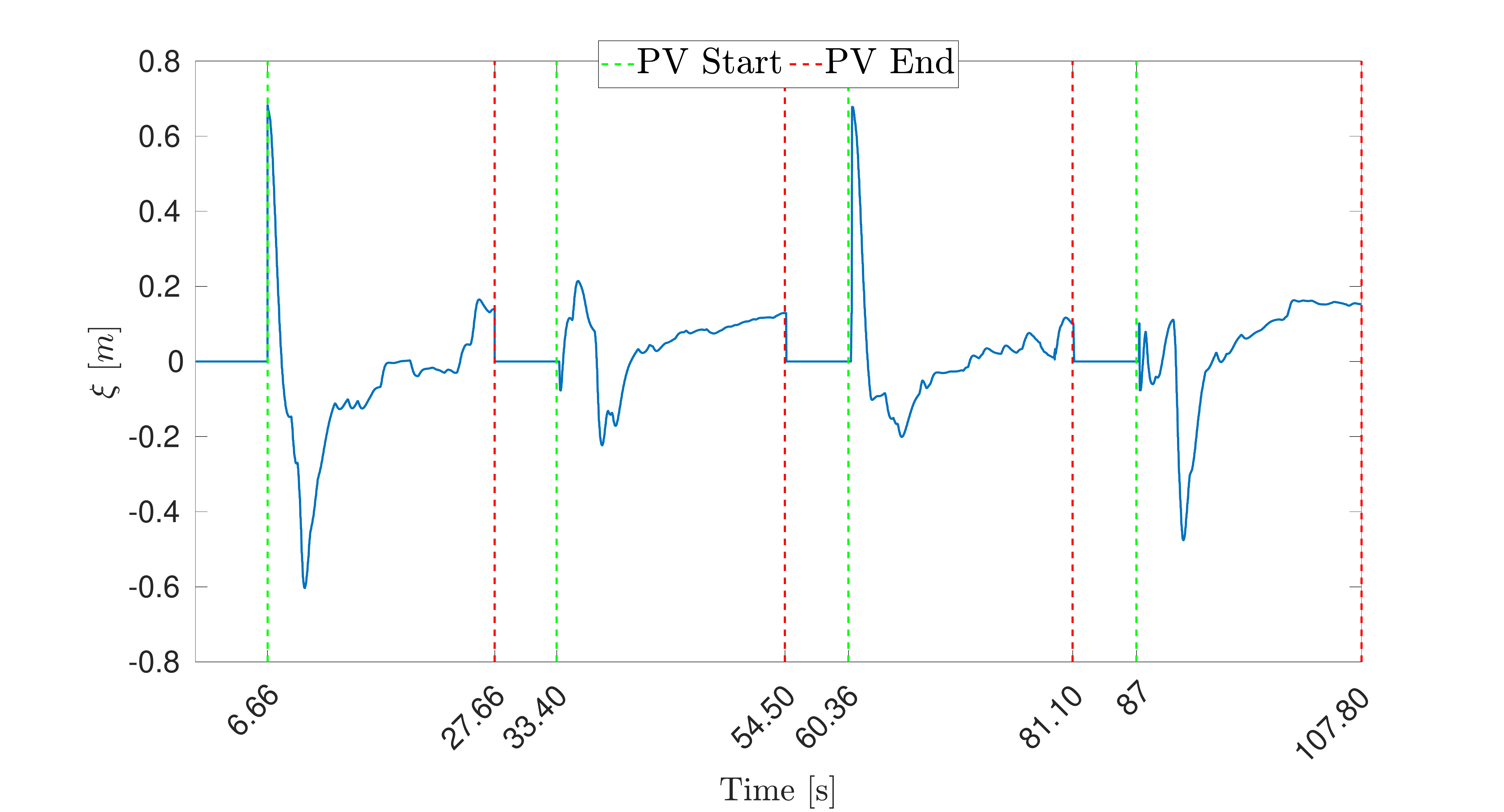}
	\caption{Navigation error $\xi$ from the real \textit{PV midline} in the presence of waypoint positioning errors (both cameras).}
	\label{fig:Figure21}
\end{figure}

Tests were performed by considering large errors in the positions of waypoints at the start and at the end of PV module rows to simulate inaccurate waypoint positions computed from Google Earth images. An example is shown in Figure \ref{fig:Figure18}, with navigation performed using both cameras: waypoints do not correspond to \textit{PV start} and \textit{PV end} of actual PV module rows since they are subject to a rototranslation error plus additional noise. 

However, observing the actual UAV's path (blue curve in Figure \ref{fig:Figure18}) shows that the UAV reached the correct \textit{PV start} on the second, third, and fourth PV row, thus compensating for large errors due to waypoint wrong placement. Figure \ref{fig:Figure19} reports errors in the \textit{PV midline} state estimated by the EKF. The control and navigation errors are comparable with those of the previous experiments: the control error in Figure \ref{fig:Figure20} has an average $\mu_e= 0.022$ m and standard deviation $\sigma_e=0.033$ m. The RMSE computed from the navigation error in Figure \ref{fig:Figure21}
is  $0.150$ m, comparable with previous experiments.

\subsection{Results in Real-World Experiments}
\label{Real Experimental Results}
Experiments were conducted in a PV plant in Predosa in Northern Italy, exploring different configurations:

\begin{itemize}
\item Subsection \ref{Navigation with Thermal Camera only} explores navigation along one PV module row using the thermal camera only.
\item Subsection \ref{Navigation with RGB Camera only} explores navigation along one PV module row using the RGB camera only.
\item Subsections \ref{Navigation with both cameras} and \ref{Navigation with both cameras along 4 rows} explore navigation along one/four PV rows using both thermal and RGB cameras.
\item Subsection \ref{Navigation with ML Thermo algorithm} explores navigation along one PV module row using an alternative, data-driven method for PV array segmentation based on a proprietary Neural Network developed by a company in Genova.
\end{itemize}

In initial tests, the UAV moved at a desired height from the ground of $15$ m and a desired speed of $1$m/s. 


\begin{figure}
	\centering
  \includegraphics[width=0.7\linewidth ,height= 5cm, keepaspectratio]{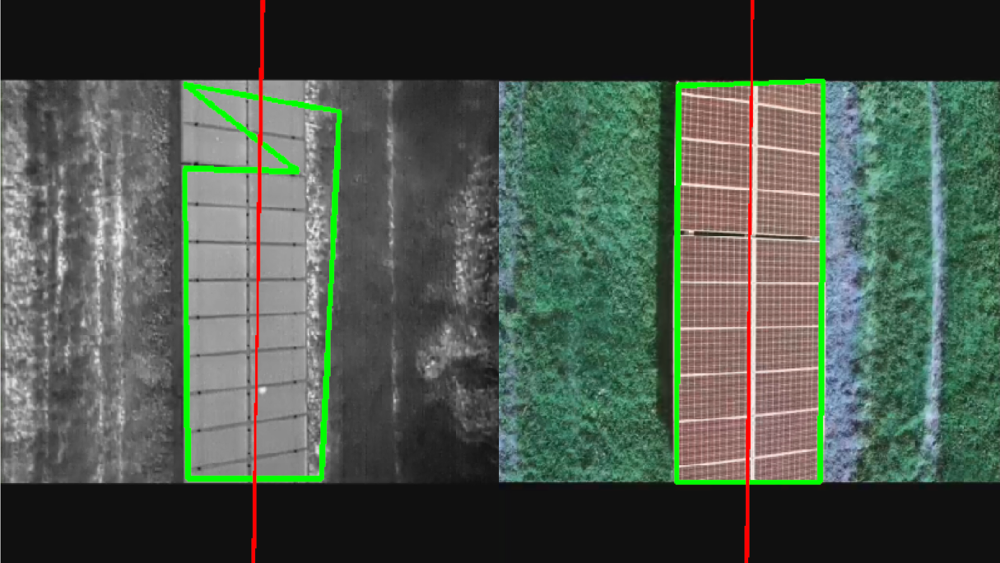}
  \caption{PV modules detection in Predosa, Italy ($15$~m from the ground).}
  \label{fig:Figure22}
\end{figure}

\subsubsection{Navigation with Thermal Camera only}

\label{Navigation with Thermal Camera only}
An example of PV module detection in Predosa, obtained using the thermal camera, is shown in Figure \ref{fig:Figure22} on the left\footnote{Image taken in the month of November at 2:55 pm, Italian local time.}. 

\begin{figure}[!t] 
\centering
  \includegraphics[width=9.5 cm, keepaspectratio]{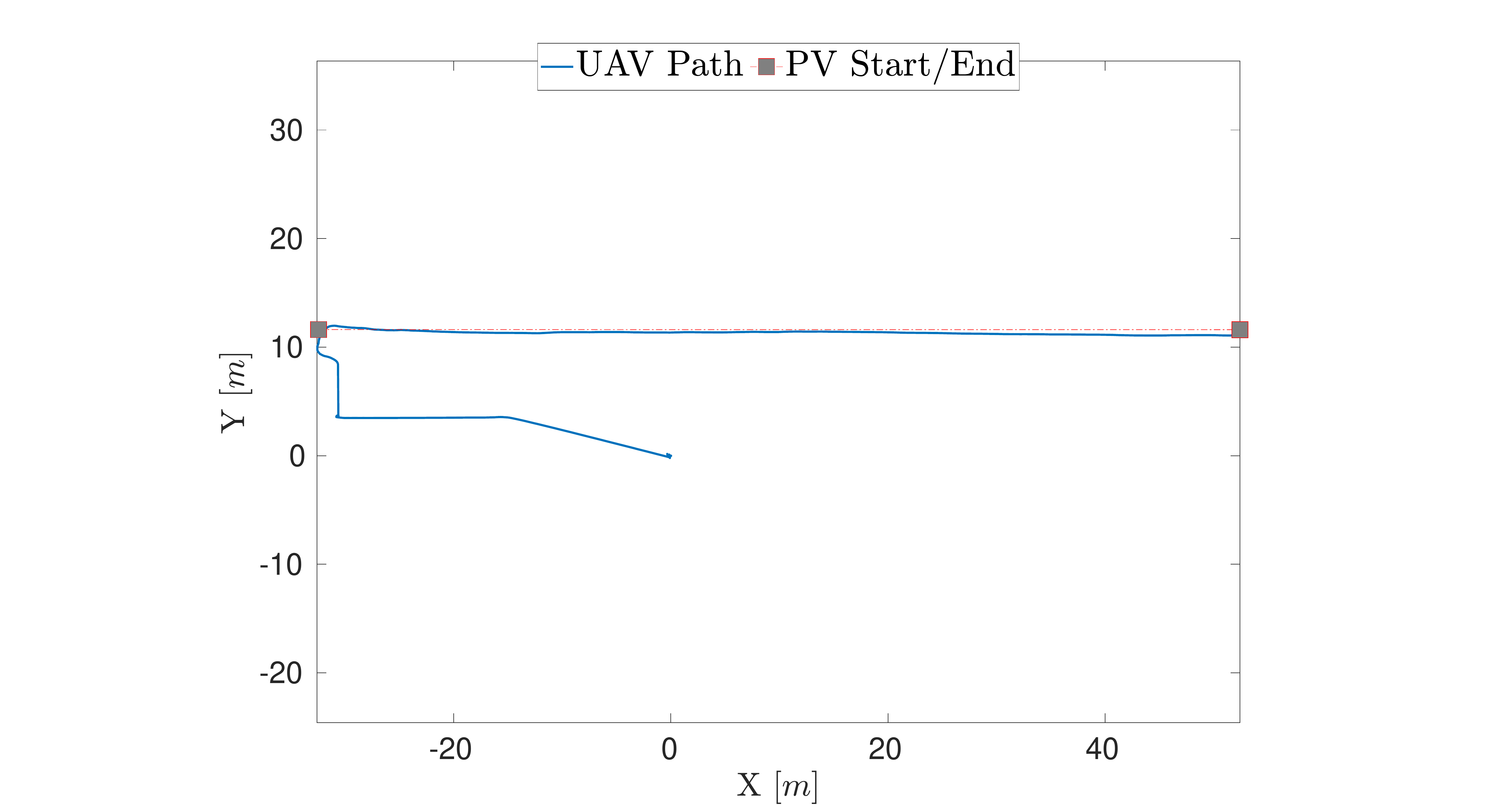}
  \caption{Projection on the XY-plane of the UAV path (thermal camera).}
    \label{fig:Figure23}
\end{figure}

\begin{figure}[t] 
  \includegraphics[width=9.5 cm, keepaspectratio]{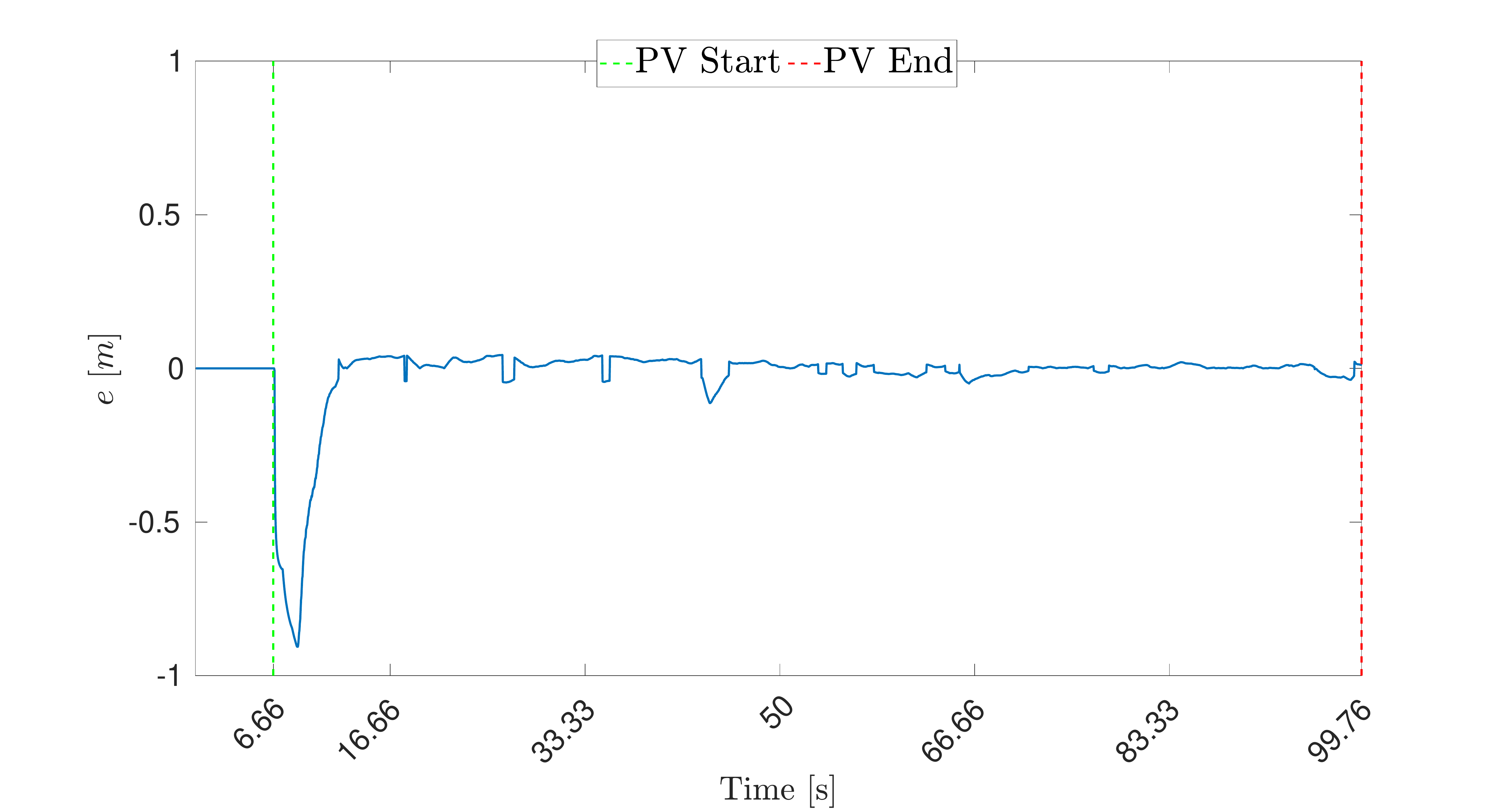}
  \caption{Control error from the estimated \textit{PV midline} (thermal camera).}
  \label{fig:Figure24}
\end{figure}

\begin{figure}[ht] 
  \includegraphics[width=9.5 cm, keepaspectratio]{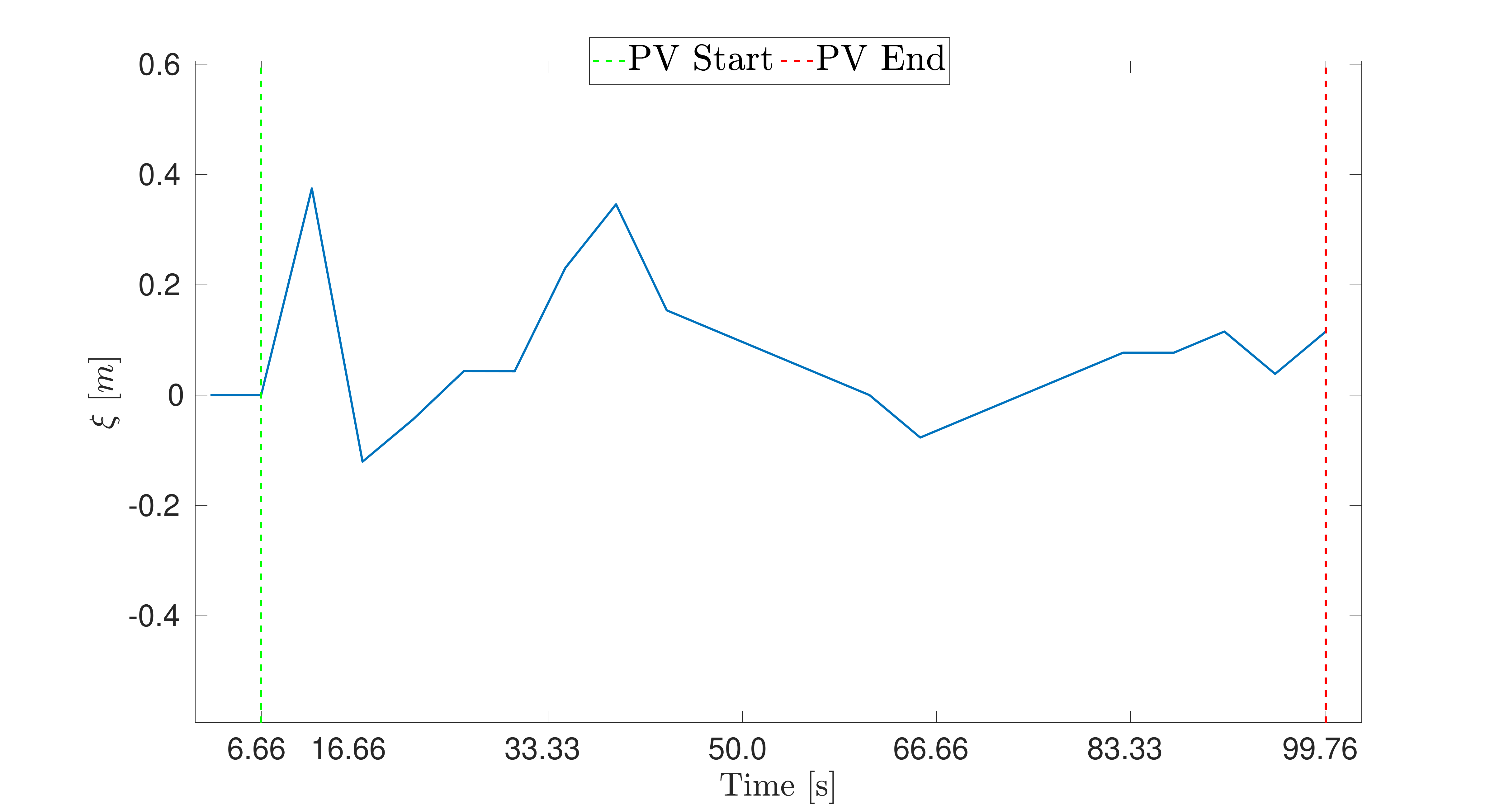}
  \caption{Navigation error from the real \textit{PV midline} (thermal camera).}
  \label{fig:Figure25}
\end{figure}

In Figure \ref{fig:Figure23}, the projection on the XY-plane of the UAV path is shown. After take-off, the UAV was manually guided to the first waypoint: in this phase, the control error $e$ and the navigation error $\xi$ are not computed, which explains why they are constantly zero in Figures \ref{fig:Figure24}-\ref{fig:Figure32}. Next, the UAV started collecting observations to estimate the parameters $\hat a^W$ and $\hat b^W$ of the \textit{PV midline}. Since the UAV does not move until the estimation error is below a given threshold at the \textit{PV start} of each row, the duration of this phase may vary in different experiments or when tracking different PV rows within the same experiment. The average control error and the RMSE associated to the navigation trajectory are computed only after the UAV has reached the speed of $1$ m/s after a transitory (i.e., the sometimes significant errors at the start of PV rows are not considered). 

Please also notice that, in real-world experiments, the UAV path and the position of the PV module row in the world, Figures \ref{fig:Figure23} and \ref{fig:Figure30}, were estimated using a GPS. However, we did not use this information for path-following but for visualization purposes only.  In all experiments, PV tracking was performed during navigation using the control method described in the previous Sections, and the navigation error $\xi$ was computed by manually inspecting the video stream acquired by onboard cameras (once again because the ground truth is not available in this case). To this end, the video stream was periodically sampled (with a period of about four seconds), and images were manually inspected to measure the distance in pixels and then in meters\footnote{The meters/pixel ratio is computed by knowing the width in meters of a PV module and measuring its corresponding width in pixels in the image.} between the center of the image and the center of the PV module row ($\xi$ tends to zero as this distance tends to zero).  


In Figure  \ref{fig:Figure24}, the control error $e$ is shown (i.e., the distance from the estimated \textit{PV midline}), with a significant initial error due to inaccurate detection of the \textit{PV midline} (which was later recovered), average  $\mu_e = 0.012$ m, and standard deviation $\sigma_e = 0.010$ m. In Figure  \ref{fig:Figure25}, the navigation error $\xi$, computed as described before, is shown (i.e., the distance from the actual \textit{PV midline}), with RMSE of $0.1571$ m.


\subsubsection{Navigation with RGB Camera only}
\label{Navigation with RGB Camera only}

An example of PV module detection using the RGB camera only is shown in Figure \ref{fig:Figure22} on the right\footnote{Image taken in the month of November at 12:50 pm Italian local time.}.

\begin{figure} 
  \includegraphics[width=9.5 cm, keepaspectratio]{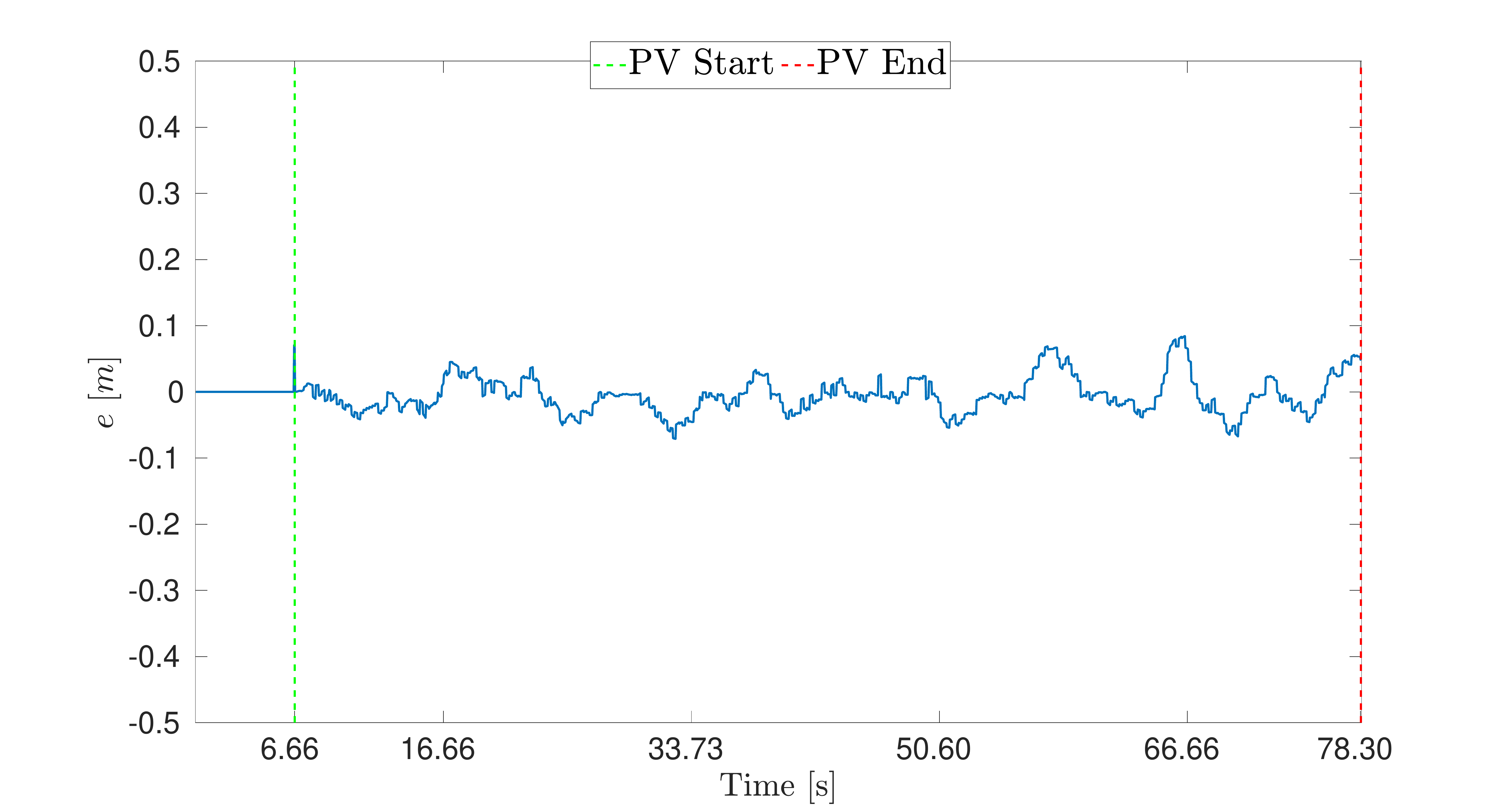}
  \caption{Control error from the estimated \textit{PV midline} (RGB camera).}
  \label{fig:Figure26}
\end{figure}
\begin{figure} 
  \includegraphics[width=9.5 cm, keepaspectratio]{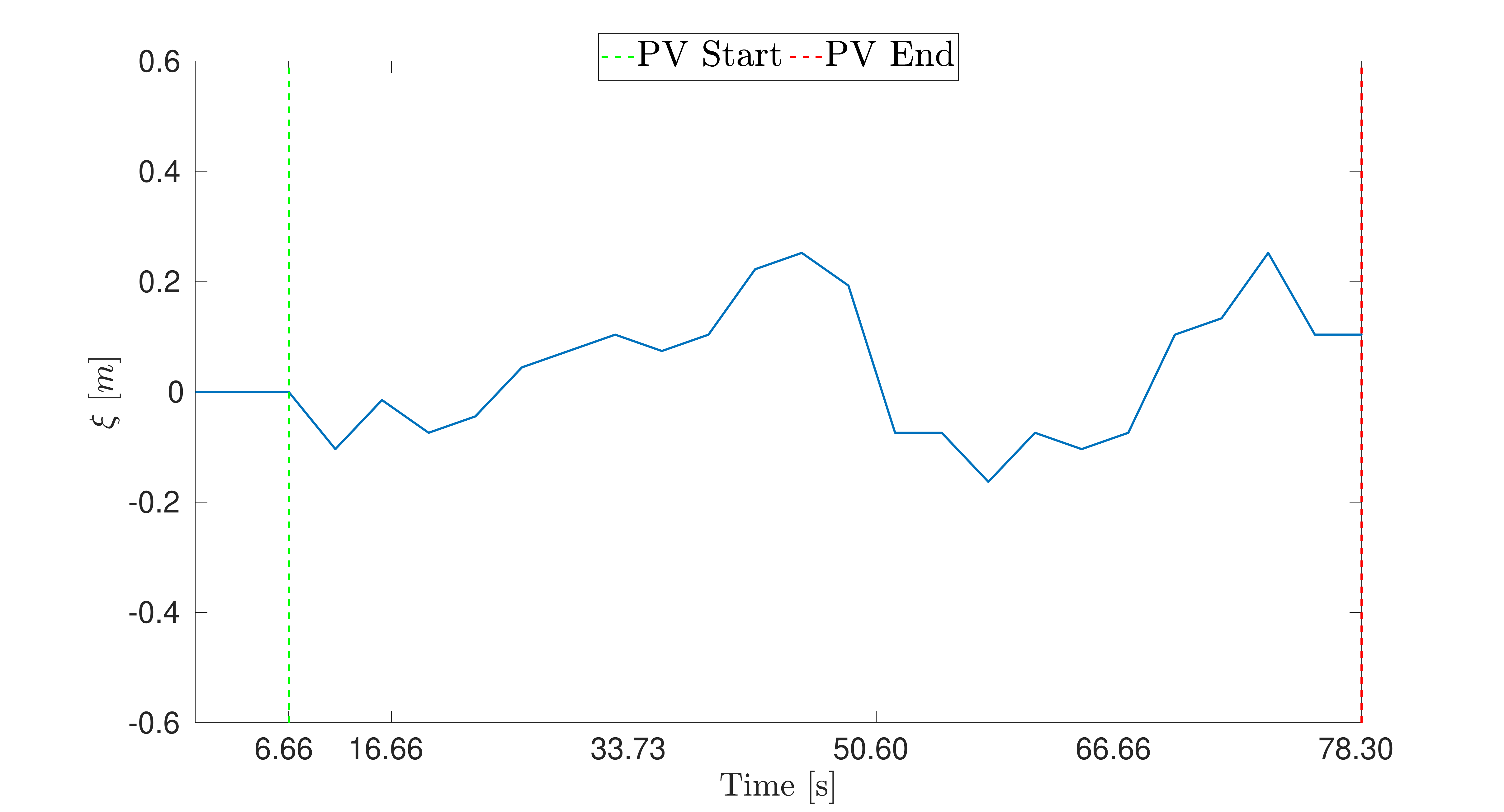}
  \caption{
Navigation error from the real \textit{PV midline} (RGB camera).}
  \label{fig:Figure27}
\end{figure}

The projection on the XY-plane of the UAV path is not reported since it would be difficult to appreciate differences with the previous test. 
In Figure  \ref{fig:Figure26}, the control error $e$ is shown, with average  $\mu_e = 0.023$ m and standard deviation $\sigma_e = 0.019$ m. In Figure \ref{fig:Figure27} the navigation error $\xi$ is shown. The RMSE is $0.1279$ m.



\subsubsection{Navigation with both cameras}
\label{Navigation with both cameras}
The same test was performed with the thermal and  RGB cameras for \textit{PV midline} estimation and navigation. 
Once again, the UAV path is not reported since it would be challenging to appreciate differences from the previous tests.


%
%

\begin{figure}[!t] 
  \includegraphics[width=9.5 cm, keepaspectratio]{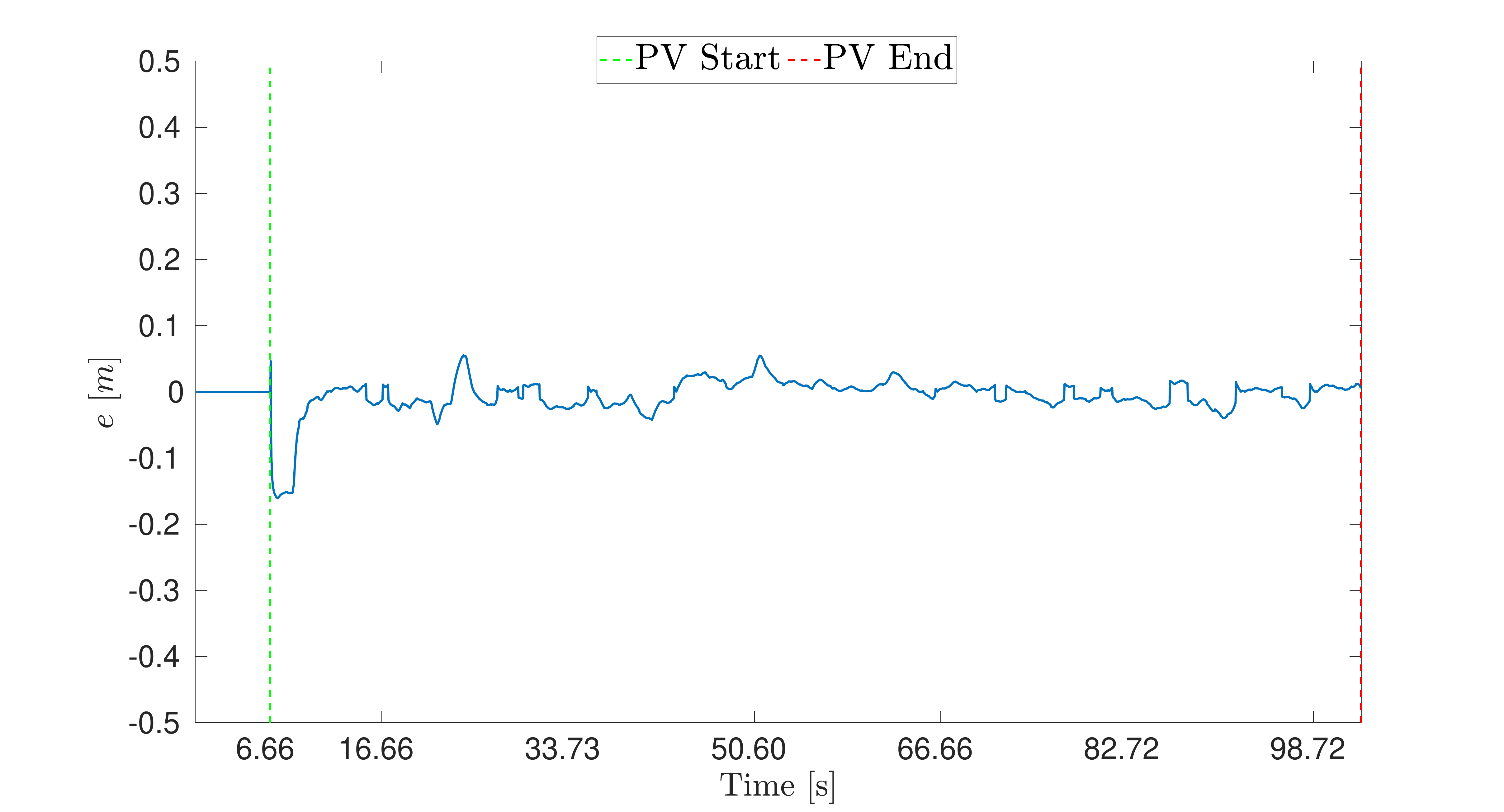}
  \caption{Control error from the estimated \textit{PV midline} (both cameras)}
  \label{fig:Figure28}
\end{figure}

\begin{figure} 
  \includegraphics[width=9.5 cm, keepaspectratio]{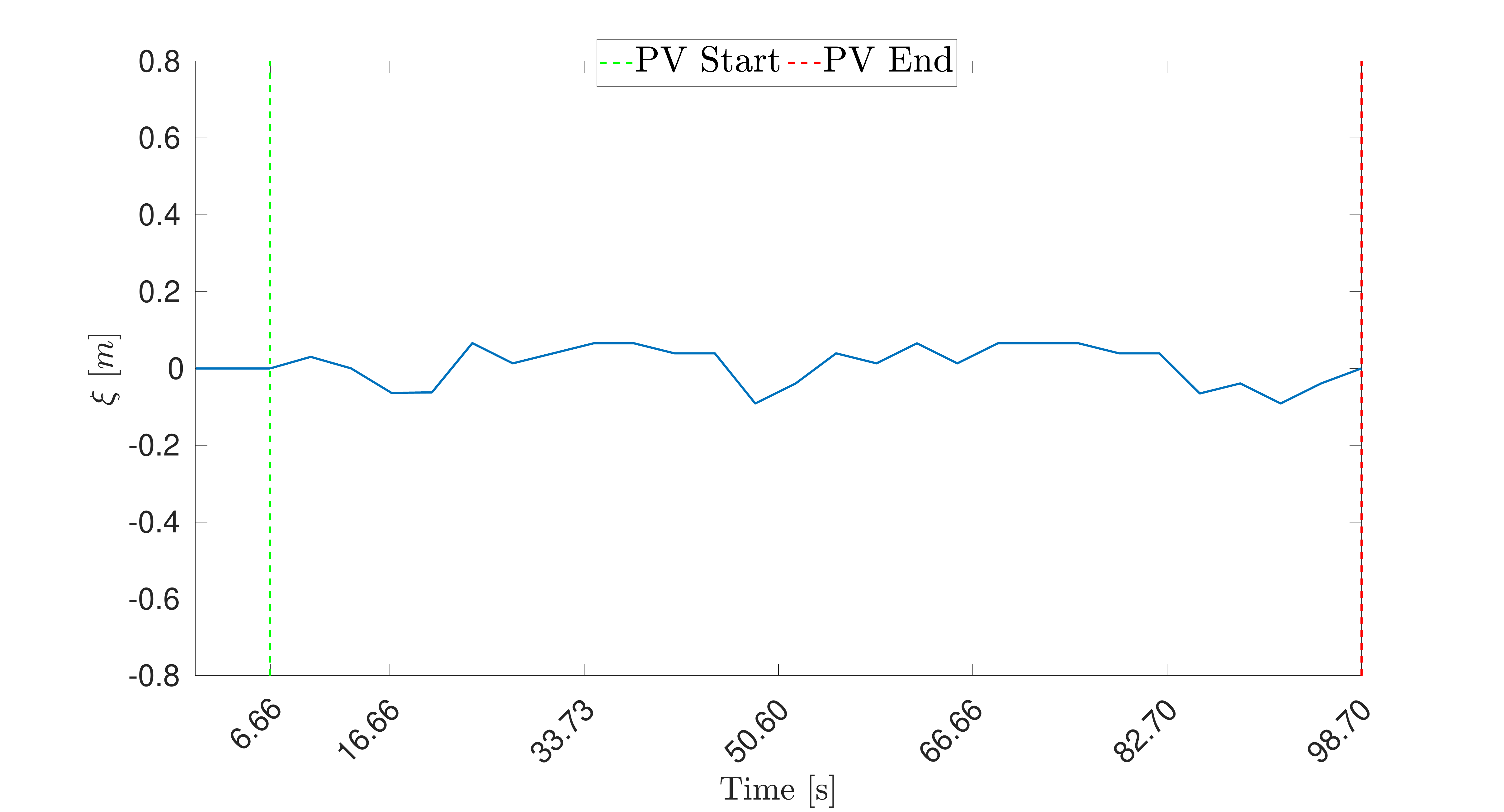}
  \caption{
Navigation error from the real \textit{PV midline} (both cameras).}
  \label{fig:Figure29}
\end{figure}

In Figure  \ref{fig:Figure28}, the control error $e$ is shown, with $\mu_e = 0.010$m  and $\sigma_e = 0.008$m. In Figure  \ref{fig:Figure29}, navigation error $\xi$ is shown, with an RMSE of
$0.0552$ m.
The average navigation error is lower than in the previous cases, as also visible in the Figure.

\subsubsection{Navigation with both cameras along 4 rows}
\label{Navigation with both cameras along 4 rows}

Additional experiments were performed, using both cameras, along a longer path involving four rows, to confirm the capability of the system to autonomously inspect PV plants. 

\begin{figure}[!t]
\centering  \includegraphics[width=9.5 cm, keepaspectratio]{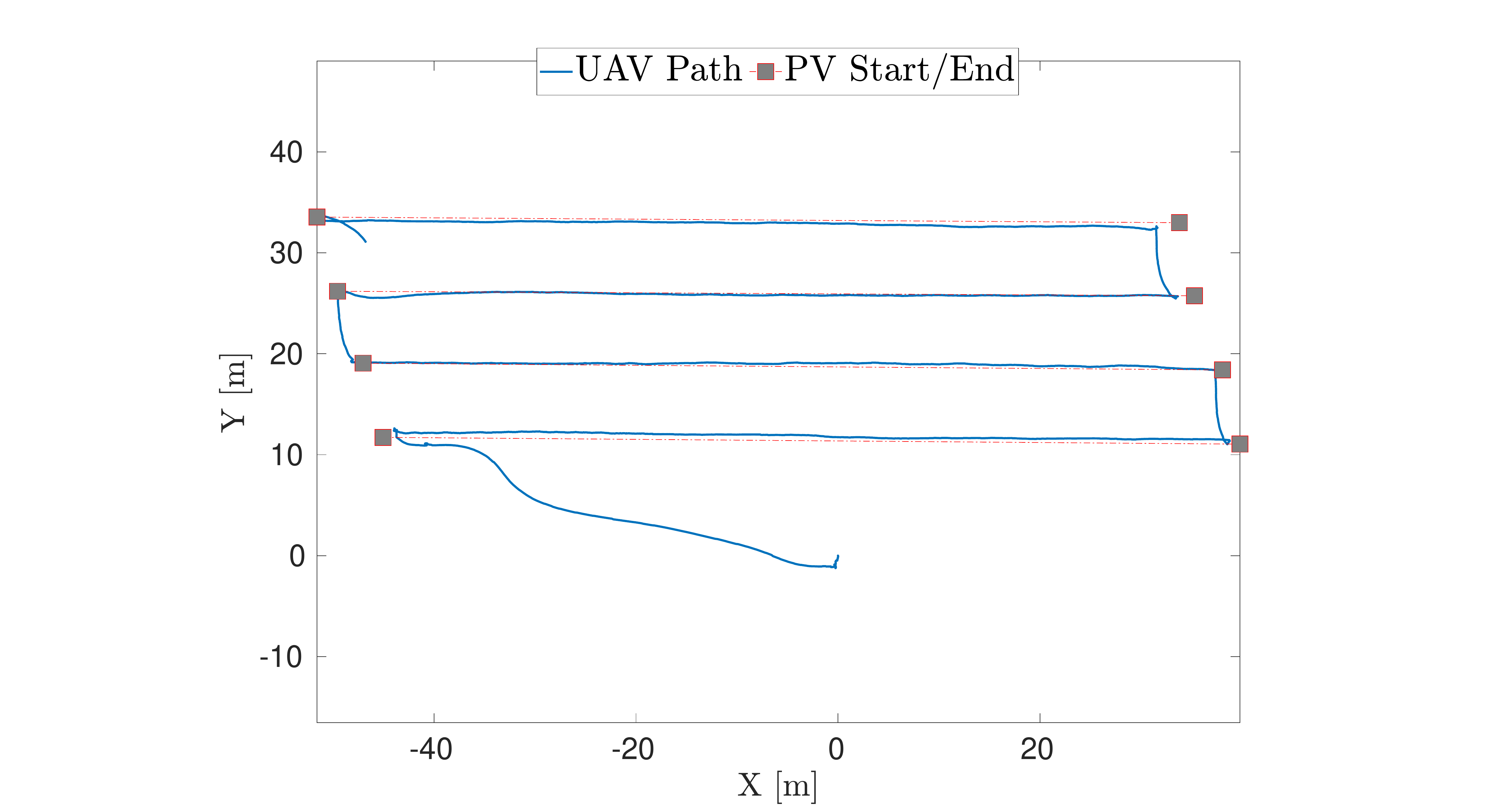}
  \caption{Projection on the XY-plane of the UAV path (both cameras).}
  \label{fig:Figure30}
\end{figure}

Figure \ref{fig:Figure30} reports the UAV path corresponding to one of these experiments with a higher speed of $1.2$ m/s\footnote{Test performed in the month of November, around $4:08$ pm CET.}. 

\begin{figure} [!t]
  \includegraphics[width=9.5 cm, keepaspectratio]{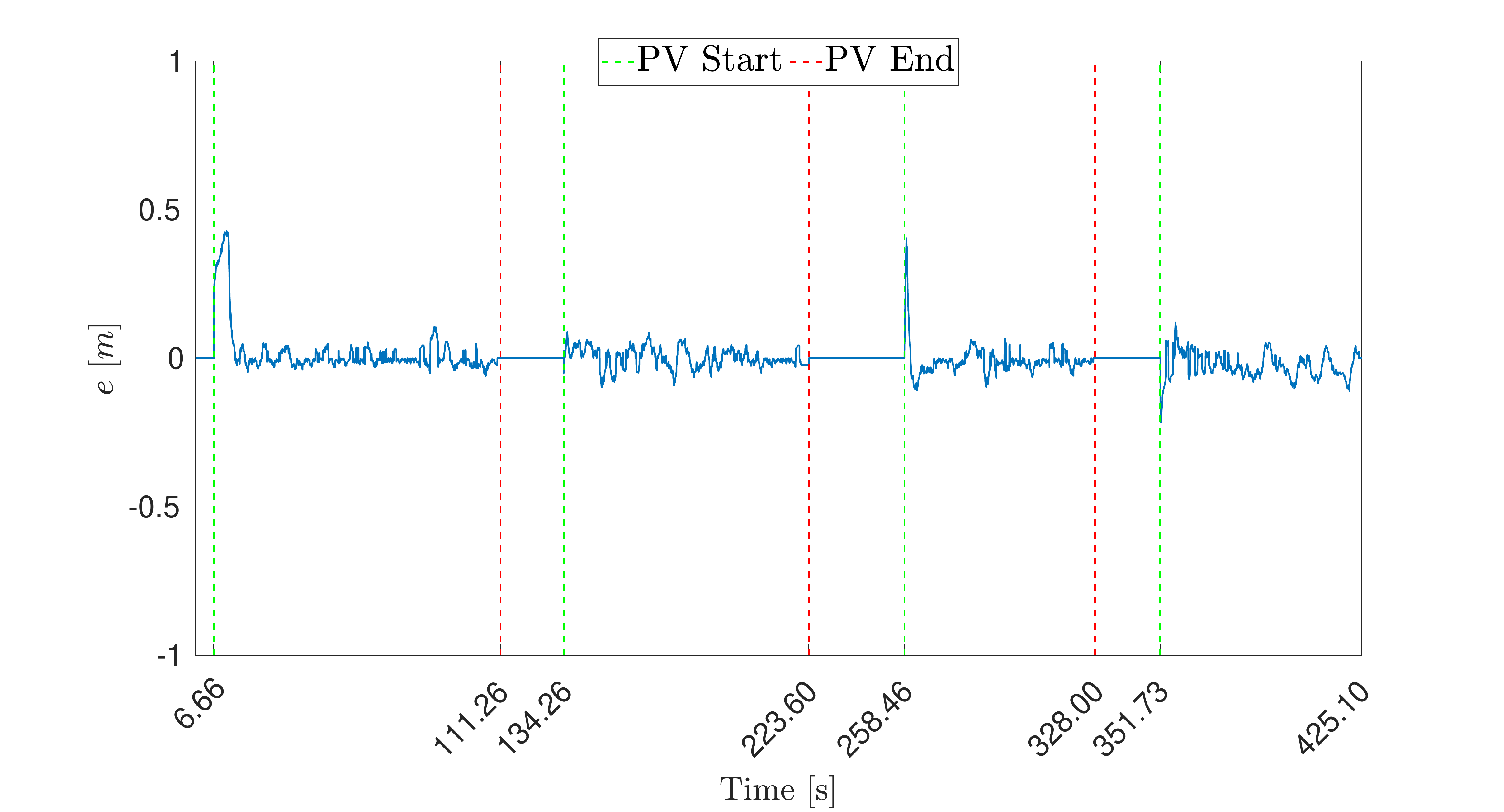}
  \caption{Control error from the estimated \textit{PV midline} (both cameras).}
  \label{fig:Figure31}
\end{figure}

\begin{figure} [!t]
  \includegraphics[width=9.5 cm, keepaspectratio]{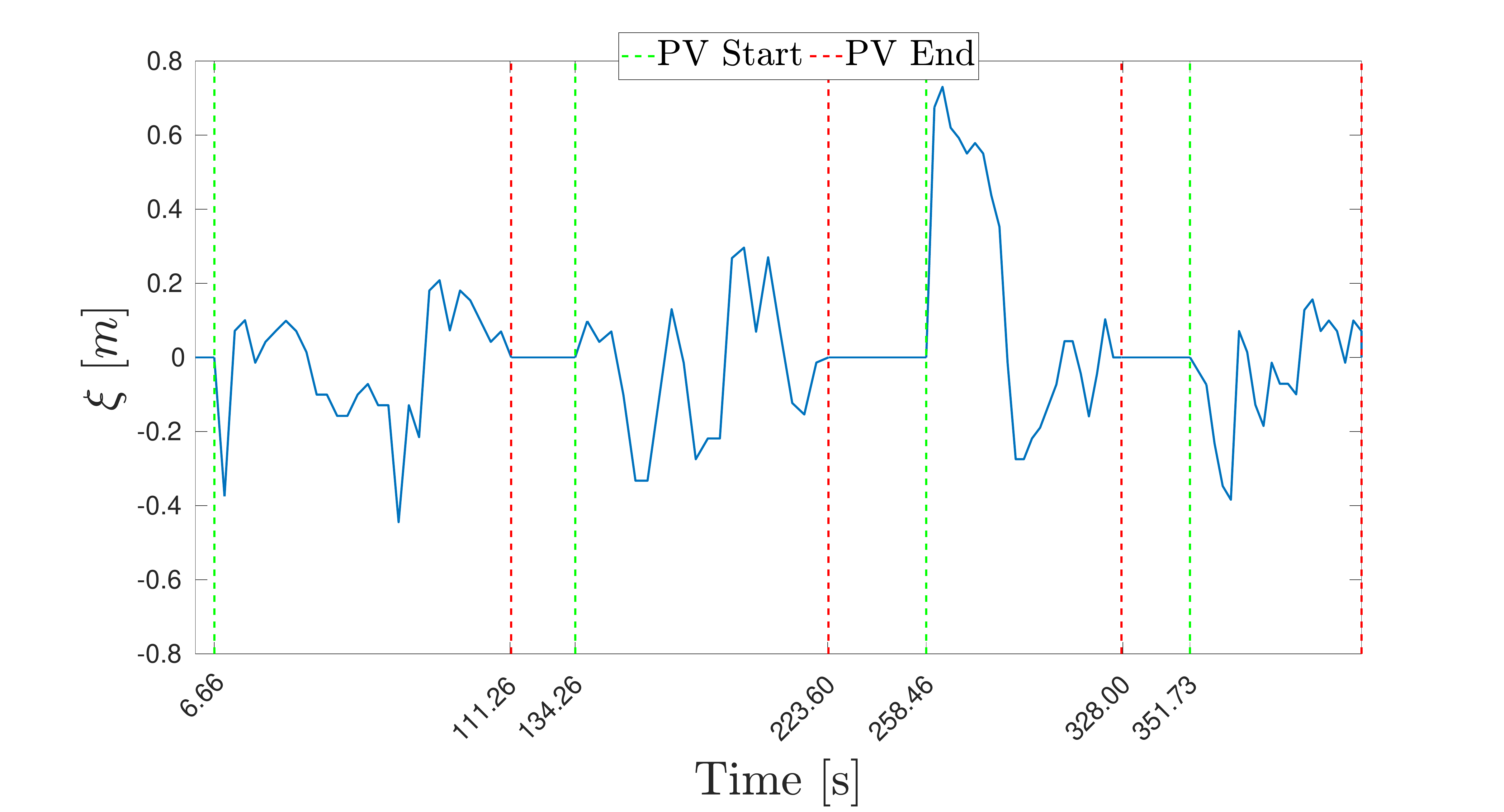}
  \caption{Navigation error from the real \textit{PV midline} (both cameras).}
  \label{fig:Figure32}
\end{figure}

Figure \ref{fig:Figure31} reports the control error: the error in the first, second, third and fourth PV rows has, respectively, average and standard deviation $\mu_e=0.021$m, $\sigma_e=0.020$m;  $\mu_e=0.019$m, $\sigma_e=0.013$m; $\mu_e=0.020$m, $\sigma_e=0.015$m; $\mu_e=0.042$m, $\sigma_e=0.024$m. Figure \ref{fig:Figure32} shows the navigation error, with the RMSE in the first, second, third, and fourth PV rows equal to $0.1594$ m, $0.1908$ m, $0.3863$ m, and $0.1549$ m. 

It can be observed that higher navigation errors can sometimes occur, especially after moving from a row to the subsequent one without PV module tracking; 
however, such error is bounded and the system is able to recover from it. 


\subsubsection{Navigation with the thermal camera using a data-driven segmentation method}
\label{Navigation with ML Thermo algorithm}

As a final test,  we also performed navigation on a single array by substituting our segmentation method with a machine learning-based (ML) algorithm for PV module detection, using only thermal images.
The ML-based segmentation algorithm is based on a Neural Network trained on a large dataset of thermal images provided by the company JPDroni, acquired during inspection flights performed at different hours of the day.  

This test had some major limitations. Since the intellectual ownership of the segmentation algorithm belongs to a third company (which also manually labeled the PV modules in each image), we were only given a ROS executable file: the company did not share any detail about the NN's inner structure. 
The ROS node was integrated into our system proposed in this article by removing the detection and clustering pipeline on thermal images described in section \ref{sec:Detection of PV arrays via thermal camera}. 

\begin{figure}[t!] 
  \includegraphics[width=9.5 cm, keepaspectratio]{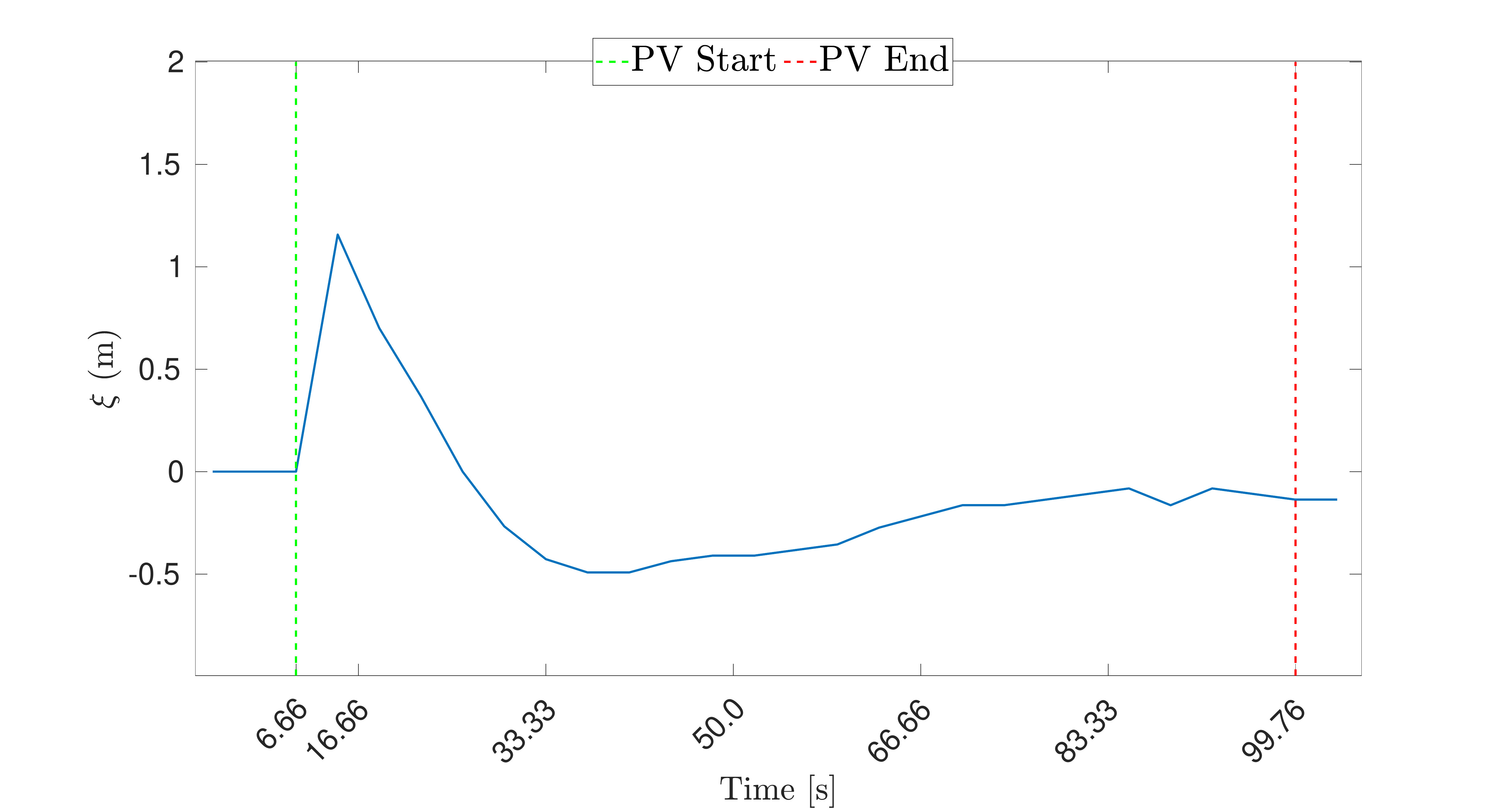}
  \caption{Navigation error from the real \textit{PV midline} (thermal camera with a NN-based perception pipeline).}
  \label{fig:Figure33}
\end{figure}

The test was conducted on the same day and PV module array as in section \ref{Navigation with Thermal Camera only} with similar environmental conditions. 
Figure \ref{fig:Figure33} reports the navigation error recorded during the flight, with an RMSE of $0.368$ m. 
As visible by comparing Figure \ref{fig:Figure33} with Figure \ref{fig:Figure25}, our model-driven approach achieves similar accuracy as this data-driven approach: the latter, however, requires several pictures of the plant in the training dataset, a requirement that our approach does not have. This limitation becomes particularly evident when changing the flying height of the UAV. If images taken at different altitudes are not part of the training set, we experimentally observed that segmentation sometimes completely fails.

\section{Conclusion}
\label{Conclusion}
This article presented a new approach for autonomous UAV inspection of a PV plant based on the detection and tracking of PV modules through thermal and RGB cameras, which is an alternative to traditional approaches based on UAV photogrammetry.  
The proposed approach is based on: 
\begin{itemize}
	\item A procedure for detecting PV modules in real-time using a thermal or an RGB camera (or both).
	\item A procedure to correct errors in the relative position of the \textit{PV midline}, initially estimated through GPS, by merging thermal and RGB data.
    \item A navigation system provided with a sequence of georeferenced waypoints defining an inspection path over the PV plant, which uses the  estimated position of the \textit{PV midline} to make the UAV move along the path with bounded navigation errors. 
	\end{itemize}

The system was tested both in a simulated and in a real PV plant with a DJI Matrice 300. Results suggest that the solution proposed meets the constraints for autonomous PV inspection. Indeed, it produces navigation errors that are sufficiently small to keep the PV modules within the camera field of view when flying at a 15m height, which allows for acquiring overlapping high-resolution thermal images for defect detection. Most importantly, navigation errors are small in the presence of errors in waypoint positions. Georeferenced waypoints computed from Google Earth images and GPS data may be affected by large biases but tend to be locally coherent. Results confirm that alternating (i) visual servoing along a PV row with (ii) GPS-based navigation when moving from one row to the subsequent one turns out to be a feasible solution.  

Tests confirmed that combining both cameras tends to reduce average errors. Additionally, we conjecture that using two sensors based on completely different principles for PV module segmentation can make the whole process more reliable when thermal or lighting conditions are suboptimal. For example, when the PV panel heat may not be clearly distinguishable from the surrounding environment (e.g., early in the morning, especially in winter) or when the sun glares can negatively affect color-based segmentation. 


In future work, we will explore the control of teams of drones \cite{Recchiuto2016}, the additional use of landmarks for more accurate navigation between subsequent PV panel rows \cite{Piaggio2001}, and finally  planning procedures to automatically extract a sequence of georeferenced waypoints from Google Earth images This task is far from being straightforward since the PV plant might have an irregular shape, be built on non-flat terrain,  and be traversed by roads or power line infrastructures. Indeed, PV module rows are not necessarily arranged in the best way to be covered following a boustrophedon path. Eventually, even if we argue that a model-based approach may offer many advantages in the considered scenario, we plan to evaluate more deeply Machine Learning techniques for PV module segmentation. This will enable fairer comparisons between model- and data-driven approaches in autonomous, UAV-based inspection of PV plants.

\end{document}